\documentclass[10pt,twocolumn,letterpaper]{article}

\usepackage[pagenumbers]{cvpr} %

\makeatletter
\@namedef{ver@everyshi.sty}{}
\makeatother

\usepackage[accsupp]{axessibility}  %
\usepackage{graphicx}
\usepackage{amsmath}
\usepackage{amssymb}
\usepackage{booktabs}

\DeclareMathOperator*{\argmin}{arg\,min}
\usepackage{tabularx}
\usepackage[table,dvipsnames]{xcolor}
\usepackage{flushend}
\usepackage{enumitem}

\usepackage[pagebackref,breaklinks,colorlinks,citecolor=blue]{hyperref}

\usepackage{tikz}
\usepackage{pgfplots}
    \usetikzlibrary{
        pgfplots.colorbrewer,
        positioning,
        calc,
        spy,
        fadings,
        patterns,
        arrows.meta
    }

    \pgfplotsset{
        cycle list/.define={my marks}{
            every mark/.append style={solid,fill=\pgfkeysvalueof{/pgfplots/mark list fill}},mark=*\\
            every mark/.append style={solid,fill=\pgfkeysvalueof{/pgfplots/mark list fill}},mark=square*\\
            every mark/.append style={solid,fill=\pgfkeysvalueof{/pgfplots/mark list fill}},mark=triangle*\\
            every mark/.append style={solid,fill=\pgfkeysvalueof{/pgfplots/mark list fill}},mark=diamond*\\
        },
    }

    \tikzstyle{closeup} = [
    opacity=1.0,          
    size=2.95cm,         
    connect spies,  %
    ]

    \usepgfplotslibrary{groupplots}
\usepackage{filecontents}
\usepackage{tabularx}

\usepackage{array} %

\usepackage[capitalize]{cleveref}
\crefname{section}{Sec.}{Secs.}
\Crefname{section}{Section}{Sections}
\crefname{subsection}{Sec.}{Secs.}
\Crefname{subsection}{Section}{Sections}
\crefname{table}{Tab.}{Tabs.}
\Crefname{table}{Table}{Tables}

\newcommand{\acro}{OCTET}
\newcommand{\bdd}{BDD100k}

\begin{document}

\title{OCTET: Object-aware Counterfactual Explanations}

\author{
  \hspace{-0.3cm} Mehdi Zemni$^1$, \hspace{0.02cm} Mickaël Chen$^1$, \hspace{0.02cm} \'Eloi Zablocki$^1$, \hspace{0.02cm} Hédi Ben-Younes$^1$, \hspace{0.02cm} Patrick Pérez$^1$, \hspace{0.02cm} Matthieu Cord$^{1,2}$ \\[0.1cm]
  $^1$ Valeo.ai, Paris, France \hspace{2cm} $^2$ Sorbonne Université, Paris, France
}

\maketitle

\begin{abstract}
Nowadays, deep vision models are being widely deployed in safety-critical applications, e.g., autonomous driving, and explainability of such models is becoming a pressing concern.
Among explanation methods, counterfactual explanations aim to find \emph{minimal} and \emph{interpretable} changes to the input image that would also change the output of the model to be explained.
Such explanations point end-users at the main factors that impact the decision of the model.
However, previous methods struggle to explain decision models trained on images with many objects, e.g., urban scenes, which are more difficult to work with but also arguably more critical to explain.
In this work, we propose to tackle this issue with an object-centric framework for counterfactual explanation generation.
Our method, inspired by recent generative modeling works, encodes the query image into a latent space that is structured in a way to ease object-level manipulations.
Doing so, it provides the end-user with control over which search directions (e.g., spatial displacement of objects, style modification, etc.) are to be explored during the counterfactual generation.
We conduct a set of experiments on counterfactual explanation benchmarks for driving scenes, and we show that our method can be adapted beyond classification, e.g., to explain semantic segmentation models.
To complete our analysis, we design and run a user study that measures the usefulness of counterfactual explanations in understanding a decision model.
Code is available at \url{https://github.com/valeoai/OCTET}.
\end{abstract}

\begin{figure}
    \centering
    \includegraphics[width=\columnwidth]{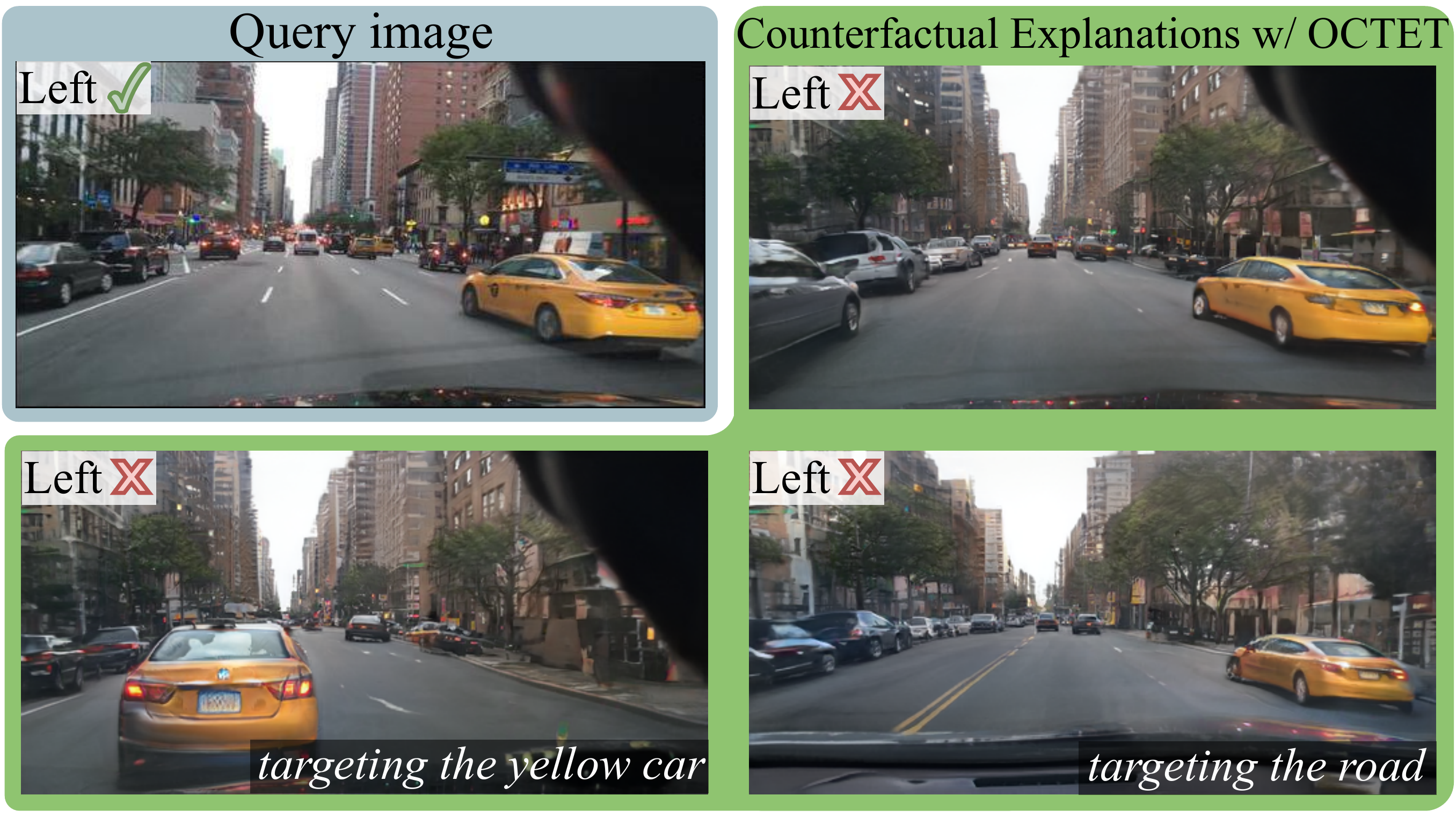}
    \caption{\textbf{Counterfactual explanations generated by \acro{}.}
    Given a classifier that predicts whether or not it is possible to go left, and a query image (\textit{top left}), \acro{} produces a counterfactual explanation where the most influential features that led to the decision are changed (\textit{top right}).
    On the bottom row, we show that \acro{} can also operate under different settings that result in different focused explanations.
    We report the prediction made by the decision model at the top left of each image.
    }
    \label{fig:teaser}
\end{figure}

\section{Introduction}
\label{sec:introduction}

Deep learning models are now being widely deployed, notably in safety-critical applications such as autonomous driving.
In such contexts, their black-box nature is a major concern, and explainability methods have been developed to improve their trustworthiness.
Among them, \emph{counterfactual explanations} have recently emerged to provide insights into a model's decision \cite{wachter2017counterfactual,counterfactual_explanations_review,explaining_classifiers_counterfactual}.
Given a decision model and an input query, a counterfactual explanation is a data point that differs \emph{minimally} but \emph{meaningfully} from the query in a way that changes the output decision of the model.
By looking at the differences between the query and the explanation, a user is able to infer --- by contrast --- which elements were essential for the model to come to its decision.
However, most counterfactual methods have only shown results for explaining classifiers trained on single-object images such as face portraits \cite{progressive-exaggeration,dive,dime,cycle-consistency-counterfactual,augustin2022diffusion_ce}.
Aside from the technical difficulties of scaling up the resolution of the images, explaining decision models trained on scenes composed of many objects also present the challenge that those decisions are often multi-factorial.
In autonomous driving, for example, most decisions have to take into account the position of all other road users, as well as the layout of the road and its markings, the traffic light and signs, the overall visibility, and many other factors.

In this paper, we present a new framework, dubbed \acro{} for Object-aware CounTerfactual ExplanaTions, to generate counterfactual examples for autonomous driving.
We leverage recent advances in unsupervised compositional generative modeling \cite{blobgan} to provide a flexible explanation method.
  Exploiting such a model as our backbone, we can assess the contribution of each object of the scene independently and look for explanations in relation to their positions, their styles, or combinations of both.

To validate our claims, extensive experiments are conducted for self-driving action decision models trained with the \bdd{} dataset \cite{bdd} and its BDD-OIA extension \cite{bdd_oia}.
\autoref{fig:teaser} shows counterfactual explanations found by \acro{}: given a query image and a visual decision system trained to assess if the ego vehicle is allowed to go left or not, \acro{} proposes a counterfactual example with cars parked on the left side that are moved closer.
When inspecting specific items (bottom row of the figure), \acro{} finds that moving the yellow car to the left, or adding a double line marking on the road, are ways to change the model's decision.
These explanations highlight that the absence of such elements on the left side of the road heavily influenced the model.

\noindent To sum up, our contributions are as follows:
\begin{itemize}[leftmargin=15pt,topsep=0pt,itemsep=0pt,parsep=0pt,partopsep=0pt]

\item We tackle the problem of building counterfactual explanations for visual decision models operating on complex compositional scenes. We specifically target autonomous driving visual scenarios.

\item Our method is also a tool to investigate the role of specific objects in a model's decision, empowering the user with control over which type of explanation to look for.%

\item We thoroughly evaluate the realism of our counterfactual images, the minimality and meaningfulness of changes, and compare against previous reference strategies. Beyond explaining classifiers, we also demonstrate the versatility of our method by addressing explanations for a segmentation network.

\item Finally, we conduct a user-centered study to assess the usefulness of our explanations in a practical case. As standard evaluation benchmarks for counterfactual explanations are lacking a concrete way to measure the interpretability of the explanations, our user-centered study is a key element to validate the presented pipeline.
\end{itemize}

\begin{figure*}
    \centering
    \includegraphics[width=\textwidth,trim={0 0 0 .31cm},clip]{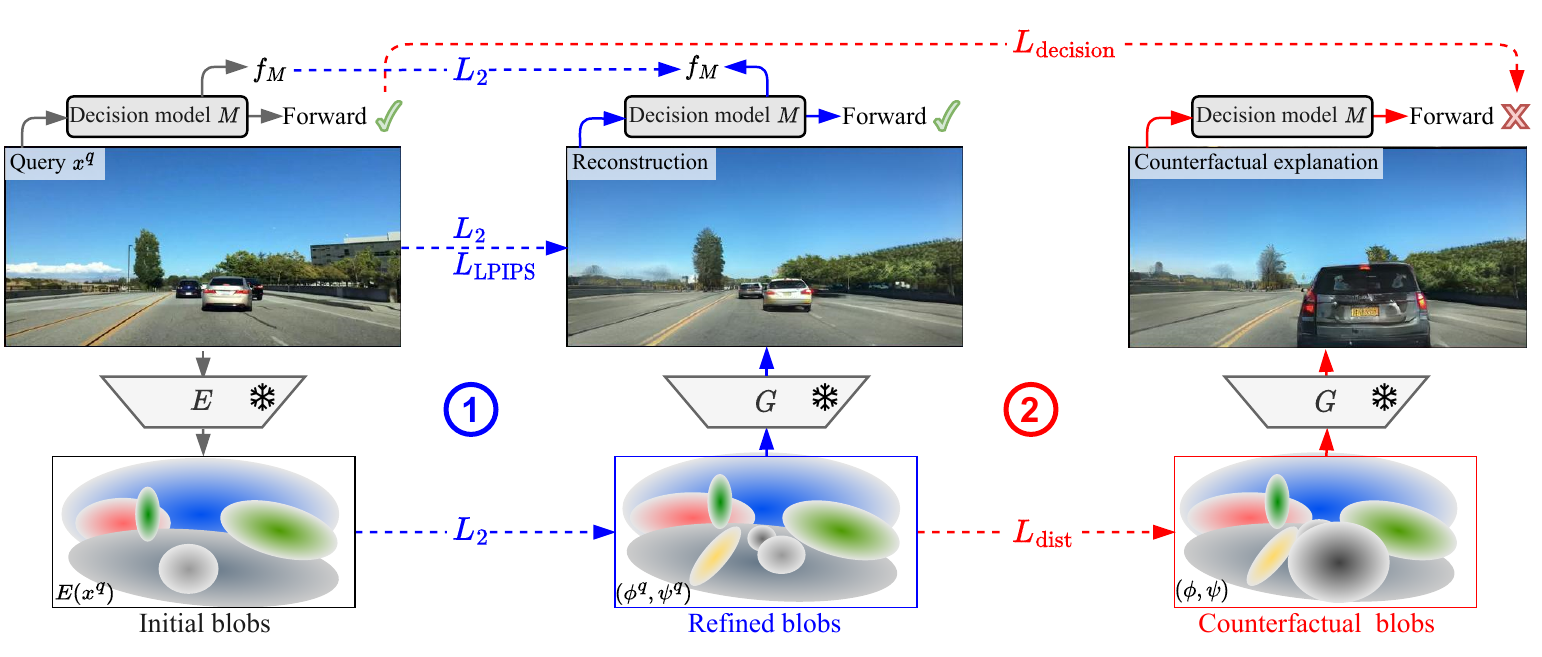}
    \caption{\textbf{Overview of \acro{}.}
    From the query image (\textit{top left}), a trained encoder $E$ provides initial blob parameters.
    The blobs are refined in optimization stage \textcolor{blue}{\raisebox{.5pt}{\textcircled{\raisebox{-.9pt} {1}}}} using \autoref{eq:linversion} to better encode the query.
    The refined blob parameters are again modified in optimization stage \textcolor{red}{\raisebox{.5pt}{\textcircled{\raisebox{-.9pt} {2}}}} using \autoref{eq:cf-latent-objective} to obtain a counterfactual example.
    The parameters of the encoder $E$ and generator $G$ are frozen during the optimization, as is $M$ the decision model to be explained.
    Losses are illustrated with dashed arrows whose directions represent the gradient flow.
    }
    \label{fig:overview}
\end{figure*}

\section{Background and Related Work}
\label{sec:rw}

\noindent\textbf{Local explanations.}\quad
The overwhelming majority of deep learning based models are designed without explainability in mind, with only a few notable exceptions \cite{interpretable_cnn,this_looks_like_that}.
This fact has prompted an interest in \emph{post-hoc} explanations of already trained models that can be useful to analyze corner cases, understand failures, and find biases \cite{lime,fel2022userstudy}.
In safety-critical applications, e.g., autonomous driving or medical imaging, explanations are especially needed for liability purposes and to foster end-user trust \cite{shen2020explain,trusthci20,driving_xai_survey,medical_xai_survey}.
Post-hoc explanations are \emph{global} if they provide a holistic view of the main decision factors driving the model \cite{soft_decision_tree,causal_learning_autoencoded_activations,stylex,discoverbiasedattr2021}, or they are \emph{local} if they target the understanding of the model behavior on a specific input \cite{gradcam,lime}.
Historically, the vast majority of local explanation methods are attribution-based: they generate saliency maps highlighting pixels or regions influencing the most the model's decision \cite{gradcam,integrated_gradients,there_and_back_again,visualbackprop,deconvnet,object_detectors_emerge_in_deep_scene,meaningful_perturbation,interpretable_fine_grained_expl,evidence-counterfactual}.
In the case of urban scenes, a saliency method would for instance point at the traffic light, or the presence of a pedestrian to explain why a driving model stops \cite{kim2017causal_attention,sauer2018affordance}.
However, saliency explanations may be misleading as they can be independent of the model at hand and merely act as edge detectors 
\cite{sanity_check_saliency}.
Besides, saliency methods are by nature restricted to show \emph{where} the important regions are and they cannot indicate \emph{what} in these regions was deemed decisive. %
For example, beyond highlighting things, it would be useful to know if the color of the light (red or green?) or the orientation of the pedestrian's body (towards or away from the car?) are taken into account in the model's decision.
Counterfactual explanations are a step in that direction.

\smallskip\noindent\textbf{Counterfactual explanations.}\quad
For a given decision model and query input, a counterfactual explanation is defined as a minimally but meaningfully modified version of the query that changes the decision of the model \cite{wachter2017counterfactual}.
The notions of `minimal' and `meaningful' are crucial here.
Adversarial attacks \cite{adversarial2014,adversarial2015,deepfool} for instance are minimal but not meaningful perturbations: they are imperceptible by design and thus not informative for the end-user \cite{ce_ae_common_grounds,connections_ce_ae,why_ce_produce_ae}.
Conversely, if not minimal, the explanation might exhibit unnecessary changes that would hinder its interpretability.
While originally developed for models operating on tabular data \cite{wachter2017counterfactual}, 
counterfactuals have been recently considered to explain deep computer vision models \cite{counterfactual-visual-explanations,progressive-exaggeration,dive,steex}. %

Seminal work \cite{grounding_visual_explanations} explored the use of natural language to produce descriptions of elements that, in case of presence, change the image classifications.
Later, retrieval-based methods aimed to explain decisions by mining the dataset for counterfactual explanations \cite{grounding_visual_explanations,counterfactual-visual-explanations,scout,heads-or-tails}.
While obtained explanations can be meaningful, the difference to the query can only be as minimal as the dataset granularity allows, limiting overall interpretability.

More recently, generator-based counterfactual explanations have been proposed \cite{dive,dime,cycle-consistency-counterfactual,augustin2022diffusion_ce,steex}. 
They exploit the power of deep generative models that map a latent space to the data distribution.
By optimizing for small manipulations in the latent space, these methods ensure for both the minimality and the meaningfulness of changes.
The properties of the generative backbone and the way the latent space is structured have a crucial impact on the resulting counterfactual explanations.
DIVE \cite{dive}, for instance, exploits the disentangled latent space of a $\beta$-TCVAE \cite{beta_tcvae} to generate diverse counterfactual explanations. However, the low visual quality of its counterfactuals hinders interpretability.
DIME \cite{dime} and DVCE \cite{augustin2022diffusion_ce} instead build on recently popularized denoising diffusion probabilistic models \cite{dhariwal2021diffusion}, achieving better-looking results but losing the more curated control on the diversity.
These works can explain classifiers working on `simple' images, e.g., face portraits \cite{celeba} or featuring a single and centered object \cite{wah2011cub,imagenet}.

Scaling up to classifiers operating over more complex images such as crowded urban scenes remains a major challenge.
To date, only STEEX \cite{steex} proposes a method able to explain decision models dealing with scenes composed of many objects.
It employs a segmentation-to-image generation backbone \cite{sean}, which helps for generating high-quality complex images but comes with two important limitations.
Firstly, it requires annotated semantic segmentation masks of images to work, which is not always available or is costly for the user to produce. %
Secondly, it imposes that the explanation keeps the same spatial and semantic layout as the query, thereby severely restricting the search space.
It is, for instance, not able to provide insights about the importance of the position of objects. %
By contrast, \acro{} does not impose a fixed spatial layout. Indeed, the structural layout of the scene is an intermediate representation of the generator, inferred from images only.
Therefore, we do not use additional annotations, and the spatial layout can be optimized by back-propagation.

\section{Object-aware Counterfactual Explanations}
\label{sec:method}

We present here \acro{}, a method to produce counterfactual examples for compositional scenes, based on a generative architecture.
We first formally describe our approach and objective in \autoref{sec:method:notations}.
Then, we instantiate the generative backbone in \autoref{sec:method:backbone} and the counterfactual objective in \autoref{sec:objective}.
Lastly, we explain how we invert the generator to recover the latent code for the query image in \autoref{sec:method:image-inversion}.
The pipeline of OCTET is shown in \autoref{fig:overview}.

\subsection{Goal and Notations}
\label{sec:method:notations}
Given a decision model $M$, a query image $x^q$, and a target decision $y \neq M(x^q)$, a counterfactual explanation is an image $x$ that is close to the query image $x^q$ but modified so that the decision $M(x)$ of the model changes to $y$.
Generative counterfactual methods, such as ours, employ a trained generator $G$ \cite{dive,dime,steex}.
We denote its latent space $\mathcal{Z}$, with $z^q \in \mathcal{Z}$ the latent code that produces the best approximation of $x^q$ when passed through $G$, \ie, so that $G(z^q) \approx x^q$.
To explain differentiable deep vision models, the definition of a counterfactual explanation stated here-above can be translated into the following optimization problem:
\begin{equation}
    \arg\!\min\limits_{z \in \mathcal{Z}} L_{\text{decision}}\big(M(G(z)), y\big) + \lambda_{\text{dist}} L_{\text{dist}}(z,z^q).
    \label{eq:cf-latent-objective}
\end{equation}
The role of the first term $L_{\text{decision}}$ is to push the decision of the model $M$ for the generated image $G(z)$ to the target class $y$;
The second term $L_{\text{dist}}$ is a distance term in the latent space of $G$, which ensures that the query $x^q$ and the explanation $G(z)$ remain similar;
$\lambda_{\text{dist}}$ is an hyper-parameter that controls the relative weight between the two terms.
With optimal $z$, $G(z)$ then recovers a counterfactual image for the query $x^q$ and the decision model $M$.
Since the choice of $L_{\text{decision}}$ is constrained by the nature of the model to explain, e.g., a classification loss if $M$ is a classifier,
we only have to specify $G$ and $L_{\text{dist}}$ to define our counterfactual method.

\subsection{Compositional generative backbone}
\label{sec:method:backbone}

In this work, we focus on explaining decision models that operate on compositional visual scenes and we look for a generative model capable of modeling such scenes. 
However, most generative models are designed for images containing centered, isolated objects, and they do not learn a latent space that can easily be disentangled into per-object representations.
Instead, we propose that our generative backbone should be object-based by design \cite{giraffe, blobgan}.
Such methods are fully differentiable and trained without supervision to generate images in a compositional fashion.

We choose to build on BlobGAN \cite{blobgan}, a compositional generative model that shows good performances and has the advantage of allowing not only edition but also insertion and removal of objects in a differentiable way. 
In BlobGAN, images are generated in two steps.
First, a \emph{layout network} maps a random Gaussian noise vector to an intermediary representation consisting of an ordered set of $K$ `\emph{blobs}'.
Each blob $k \in \{1\ldots K\}$ is defined by spatial parameters $\phi_k$ and a style feature vector $\psi_k$.
The spatial parameters $\phi_k$ consist of five values that are its center coordinates, scale, aspect ratio, and rotation angle.
In practice, the layout network is a neural network that outputs vectors $\phi_k$ and $\psi_k$ for each blob. 
Then, in the second stage, the \emph{generator network} is tasked to transform the blob parameters into an image.
A rendering layer draws the blobs on a canvas as ellipses whose shapes and positions are given directly by $\phi_k$.
The output of the rendering layer is a map that associates to each 2D spatial location a feature vector computed from the style vectors of the blobs present at that position.
This feature map is transformed by convolutional layers into an image that roughly follows the spatial organization of the canvas.

The layout network and the generator network are trained end-to-end against a discriminator with standard GAN losses only.
In particular, we do not use any annotations for the number, size, class, or location of objects.
They are discovered by the generator during training.
Note also that scale parameters can take non-positive values.
In that case, the object is considered absent and the corresponding blob is not rendered on the canvas in the generator network.

\subsection{Setting $L_\text{dist}$ for compositional scenes}
\label{sec:objective}

Using a compositional generative backbone allows us to design a meaningful, disentangled, per-object distance for images.
Indeed, to be interpretable with respect to the query image, a counterfactual explanation should display sparse changes in terms of the number of moved, added, removed, or modified objects.
Therefore, we design $L_\text{dist}$ such that it is split into per-object distances where each object can be addressed independently.
Moreover,
we propose that modifications of the spatial position of an object should be orthogonal to those made to its shape or colors.
Since in a well-trained BlobGAN each blob is associated with a distinct element of the scene (see Appendix), such a distance can be naturally defined in the space of blob parameters.

Formally, let us denote $\phi$ and $\psi$ the concatenation across all blobs $k \in\{1\dots K\}$ of spatial parameters $\phi_k$ and style vectors $\psi_k$, and $z = (\phi, \psi)$ the complete latent representation that the generator network $G$ can decode into an image. 
Accordingly, the latent code for the query image $x^q$ reads $z^q = (\phi^q, \psi^q)$. 
Then, \acro{} implements $L_{\text{dist}}$ as follows:
\begin{equation}
    L_{\text{dist}}(z, z^q) = \\ \sum\limits_{k=1}^{K} \|\phi_k - \phi_k^q\|_1 + \|\psi_k - \psi_k^q\|_1.
    \label{eq:ldist-objects}
\end{equation}
We use the $L_1$ distance here to promote sparsity in the number of modified blobs and the type of edits applied to each blob. 
If the user wishes to inspect the role of the style of objects, their spatial positions, or both, with a counterfactual explanation, she can instead optimize \autoref{eq:ldist-objects} only on the corresponding features of the blobs. 

In order to specify \autoref{eq:ldist-objects}, and thus the final counterfactual objective (\autoref{eq:cf-latent-objective}), it remains to be solved how to obtain the blob-wise latent codes $\phi_k^q$ and $\psi_k^q$ associated with the query image $x^q$.
We discuss this matter in the next section.

\subsection{Inverting the query image}
\label{sec:method:image-inversion}

\begin{figure*}
    \centering
    \includegraphics[width=\textwidth]{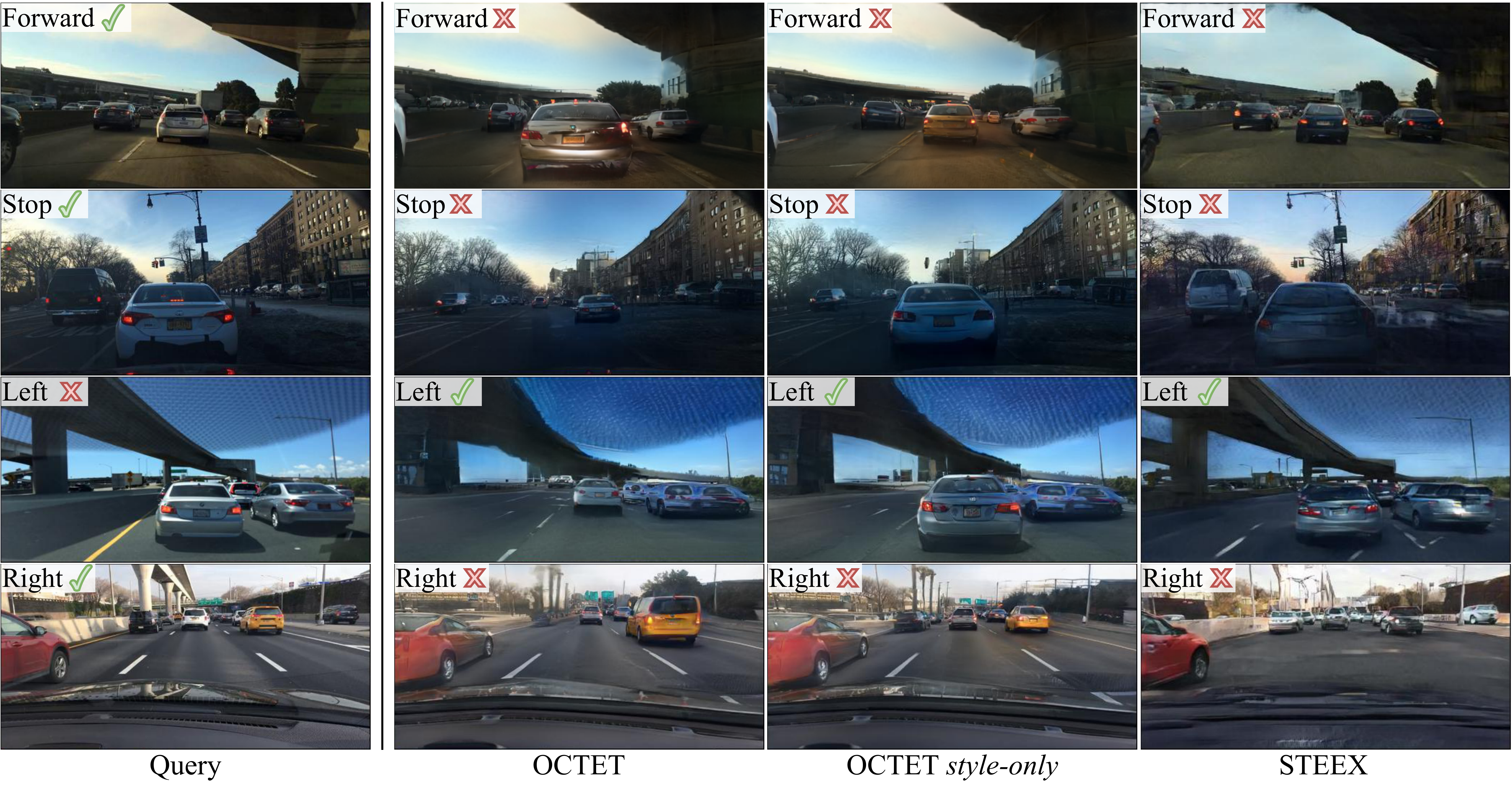}
\vspace{-0.6cm}
    \caption{
    \textbf{Counterfactual explanations on driving scenes}. Explanations are generated for binary classifiers predicting the capacity of the ego-car to respectively Move forward, Stop, turn Left, or Right.
    We indicate at the top left of each image the output decision of the model for that image.
    For each row, counterfactuals target the opposite decision to that obtained for the query image.
    \acro{} finds interpretable, sparse, and meaningful semantic and spatial modifications to the query image.
    \acro{} \emph{style-only} only optimizes on the style features, it is analog to STEEX \cite{steex} that can only search for changes in object appearance to flip the decision.
}
    \label{fig:qualitative}
\end{figure*}

Obtaining the latent code corresponding to the query image, a process called \emph{inversion}, is not straightforward when using a GAN-based generator.
To produce a counterfactual example, such inversion needs to find latent codes $(\phi^q, \psi^q)$ so that $G(\phi^q, \psi^q)$ recovers $x^q$.
For the downstream counterfactual optimization to make sense, the decision of the model for the query image should be preserved for the reconstructed image, \ie, $M(G(\phi^q, \psi^q)) = M(x^q)$.

To achieve those objectives, we follow best practices for image inversion \cite{Image2StyleGAN, Image2StyleGAN++, idinvert, e4e, psp}.
We first learn an encoder $E$ to predict blob parameters %
from input images.
Generating an image $G(E(x^q))$ with parameters predicted this way yields a coarse reconstruction of $x^q$. To improve the inversion, we start from predicted blob parameters $E(x^q)$ and further optimize them with the following objective:
\begin{equation}
\begin{split}
    \phi^q, \psi^q = \argmin\limits_{\phi,\psi} ~ & L_\text{LPIPS}(G(\phi, \psi), x^q)) \hspace{-0.07cm} + \hspace{-0.07cm} L_{2}(G(\phi, \psi), x^q) \\[-0.25cm]
    & \hspace{1cm} + L_2(f_M(x^q), f_M(G(\phi, \psi)))\\
    & \hspace{1cm} + L_{2}((\phi, \psi), E(x^q)).
\end{split}
\raisetag{0.5cm}
\label{eq:linversion}
\end{equation}
The first two terms ensure that the image $G(\phi,\psi)$ generated from the blob parameters reconstructs the query $x^q$ using a perceptual $L_\text{LPIPS}$ loss and the $L_2$ loss.
The third term aims to preserve the features that are important to the decision model $M$ by adding a $L_2$ loss on intermediate features $f_M(x)$ of $M$.
Finally, the last term encourages the parameters to stay close to the reasonably good solution given by the encoder.
This inversion method concludes the specification of the optimization problem of \autoref{eq:cf-latent-objective} in \acro{} and of the pipeline depicted in \autoref{fig:overview}.
More details on the inversion and the training of the encoder are in the appendix.

\section{Experiments}
\label{sec:experiments}

\subsection{Experimental protocol and training details}
\label{sec:experiments:protocol-details}

\paragraph{Data and decision model.}
Our method is designed to build counterfactual examples to explain decision models trained on compositional images.
To evaluate our method, we follow previous work \cite{steex} and use the \bdd{} dataset \cite{bdd}, which contains 100k images of diverse driving scenes that we resize to a resolution of 512x256.
In addition, we also use BDD-OIA \cite{bdd_oia}, a 20K scene extension of \bdd{} that is annotated with 4 binary attributes labeled \textit{Move Forward}, \textit{Stop}, \textit{Turn Left}, and \textit{Turn Right}.
Each label is defined as the capacity in the given situation of the ego-vehicle to perform the corresponding action.
Note that by this definition, being allowed to \textit{Stop} and to \textit{Move Forward} are not mutually exclusive.
The decision models being explained in the experiments are multi-label binary DenseNet121 \cite{densenet} classifiers trained on BDD-OIA with the four aforementioned labels.

\smallskip\noindent\textbf{Baseline.}\quad
We compare our method against STEEX \cite{steex}, the only counterfactual explanation method that performs well on compositional images like driving scenes.
Note that in STEEX, the conservation of the semantic layout is hard-coded in the architecture: it is less flexible as an explanation method than ours.
Moreover, the authors use semantic segmentation annotations of \bdd{} to train their generative backbone, while we do not need any additional annotation.
We use the pre-trained model from the official release.

\smallskip\noindent\textbf{Technical setup.}\quad
For our method, we train the generative backbone (BlobGAN) on the training set of \bdd{}, with a number $K=40$ of blobs. 
We adapt the architecture of BlobGAN to generate rectangular 512$\times$256 images, and we decrease the size of the feature vectors
of the BlobGAN to compensate for the increased number of blobs. 
Throughout our experiments, we evaluate two main variants of our method: \acro{} where both spatial and style features are optimized altogether and 
\acro{} \emph{style-only} where only the style vector $\psi$ is optimized while the spatial vector $\phi$ is left equal to $\phi^q$.
This setting only assesses the contribution of style features, and is most similar to the STEEX baseline \cite{steex} that finds explanations while conserving the semantic layout of the query. 
More details in the appendix.

\subsection{Quality of the explanations}
\label{sec:experiments:quality}

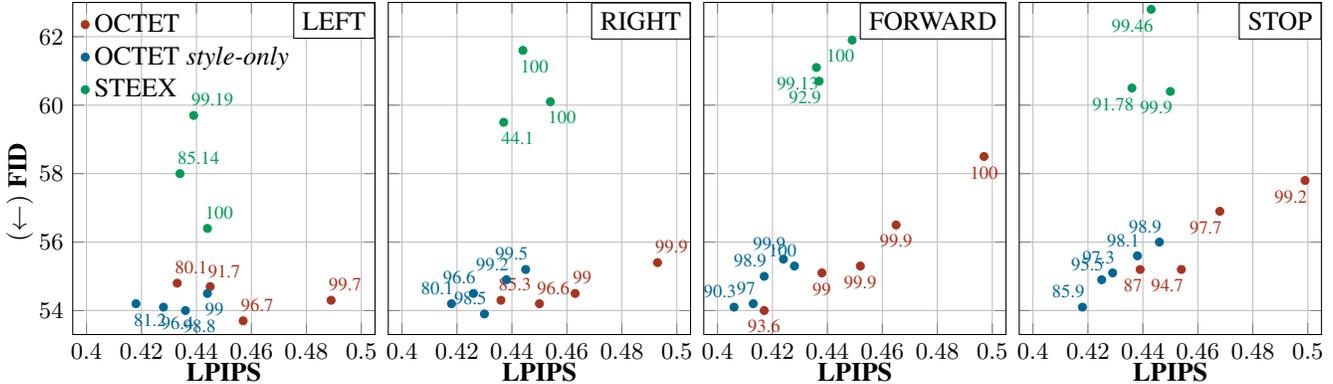
\begin{figure*}
\begin{flushright}
\begin{tikzpicture}[trim axis left, trim axis right]
\begin{axis}[%
    title=LEFT,
    title style={at={(1,1)},
                   anchor=north east,
                   draw=black,fill=white,
                   yshift=-6pt,},
    xmin=.395, xmax=.505,
    ymin=53.3, ymax=63,
    width=5.6cm, height=6.cm,
    grid=both,
    xtick distance=.02,
    ytick distance=2,
    xlabel=\textbf{LPIPS},
    ylabel=$(\leftarrow)~$\textbf{FID},
    ylabel style={xshift=-9pt, yshift=-15pt},   
    xlabel style={yshift=8pt,},
    tick label style = {font=\small},
    scatter, mark=*, mark size=1.5pt,
    scatter/classes={a={Mahogany}, b={MidnightBlue}, c={ForestGreen}},
    scatter src=explicit symbolic,
    nodes near coords style = {font=\scriptsize, inner xsep=-1pt},
    nodes near coords*={\Label},
    visualization depends on={value \thisrow{sr} \as \Label},
    legend cell align={left},
    legend style={
            at={(0,1)},
            anchor=north west,
            draw=none, %
            fill=none, %
        },
    ]
    \addplot[
    only marks,
    nodes near coords style = {text=Mahogany, anchor=south west},
    ] 
    table[meta=class, x index=0] {
    lpips fid class sr
    0.547 54.6 a 100
    0.489 54.3 a 99.7
    0.457 53.7 a 96.7
    0.445 54.7 a 91.7
    0.433 54.8 a 80.1
    };
    \addlegendentry{\acro{}}
    \addplot[
    only marks,
    nodes near coords style = {text=MidnightBlue, anchor=north west},
    ] 
    table[meta=class, x index=0] {
    lpips fid class sr
    0.444 54.5 b 99
    0.436 54 b 98.8
    0.428 54.1 b 96.4
    0.418 54.2 b 81.2
    };
    \addlegendentry{\acro{} \emph{style-only}}
    \addplot[
    only marks,
    nodes near coords style = {text=ForestGreen, anchor=south west},
    ] 
    table[meta=class, x index=0] {
    lpips fid class sr
    0.444 56.4 c 100
    0.439 59.7 c 99.19
    0.434 58 c 85.14
    };
    \addlegendentry{STEEX}%
\end{axis}
\end{tikzpicture}
\begin{tikzpicture}[trim axis left, trim axis right]
\begin{axis}[%
    title=RIGHT,
    title style={at={(1,1)},
                   anchor=north east,
                   draw=black,fill=white,
                   yshift=-6pt,},
    xmin=.395, xmax=.505,
    ymin=53.3, ymax=63,
    width=5.6cm, height=6.cm,
    grid=both,
    xtick distance=.02,
    ytick distance=2,
    xlabel=\textbf{LPIPS},
    ylabel style={xshift=-9pt, yshift=-15pt},
    xlabel style={yshift=8pt,},, %
    yticklabels={,,},
    tick label style = {font=\small},
    scatter, mark=*, mark size=1.5pt,
    scatter/classes={a={Mahogany}, b={MidnightBlue}, c={ForestGreen}},
    scatter src=explicit symbolic,
    nodes near coords style = {font=\scriptsize, inner xsep=-1pt},
    nodes near coords*={\Label},
    visualization depends on={value \thisrow{sr} \as \Label},
    legend cell align={left},
    legend style={
            at={(0,1)},
            anchor=north west,
            draw=none, %
            fill=none, %
        },
    ]
    \addplot[
    only marks,
    nodes near coords style = {text=Mahogany, anchor=south west},
    ] 
    table[meta=class, x index=0] {
    lpips fid class sr
    0.546 59.1 a 100
    0.493 55.4 a 99.9
    0.463 54.5 a 99
    0.45 54.2 a 96.6
    0.436 54.3 a 85.3
    };
    \addplot[
    only marks,
    nodes near coords style = {text=MidnightBlue, anchor=south east},
    ] 
    table[meta=class, x index=0] {
    lpips fid class sr
    0.445 55.2 b 99.5
    0.438 54.9 b 99.2
    0.430 53.9 b 98.5
    0.426 54.5 b 96.6
    0.418 54.2 b 80.1
    };
    \addplot[
    only marks,
    nodes near coords style = {text=ForestGreen, anchor=north west},
    ] 
    table[meta=class, x index=0] {
    lpips fid class sr
    0.454 60.1 c 100
    0.444 61.6 c 100
    0.437 59.5 c 44.1
    };
\end{axis}
\end{tikzpicture}
\begin{tikzpicture}[trim axis left, trim axis right]
\begin{axis}[%
    title=FORWARD,
    title style={at={(1,1)},
                   anchor=north east,
                   draw=black,fill=white,
                   yshift=-6pt,},
    xmin=.395, xmax=.505,
    ymin=53.3, ymax=63,
    width=5.6cm, height=6.cm,
    grid=both,
    xtick distance=.02,
    ytick distance=2,
    xlabel=\textbf{LPIPS},
    ylabel style={xshift=-9pt, yshift=-15pt},
    xlabel style={yshift=8pt,},
    yticklabels={,,},
    tick label style = {font=\small},
    scatter, mark=*, mark size=1.5pt,
    scatter/classes={a={Mahogany}, b={MidnightBlue}, c={ForestGreen}},
    scatter src=explicit symbolic,
    nodes near coords style = {font=\scriptsize, inner xsep=-1pt},
    nodes near coords*={\Label},
    visualization depends on={value \thisrow{sr} \as \Label},
    ]
    \addplot[
    only marks,
    nodes near coords style = {text=Mahogany, anchor=north},
    ] 
    table[meta=class, x index=0] {
    lpips fid class sr
    0.497 58.5 a 100
    0.465 56.5 a 99.9
    0.452 55.3 a 99.9
    0.438 55.1 a 99
    0.417 54 a 93.6 
    };
    \addplot[
    only marks,
    nodes near coords style = {text=MidnightBlue, anchor=south east},
    ] 
    table[meta=class, x index=0] {
    lpips fid class sr
    0.428 55.3 b 100
    0.424 55.5 b 99.9
    0.417 55 b 98.9
    0.413 54.2 b 97
    0.406 54.1 b 90.3
    };
    \addplot[
    only marks,
    nodes near coords style = {text=ForestGreen, anchor=north east},
    ]
    table[meta=class, x index=0] {
    lpips fid class sr
    0.449 61.9 c 100
    0.442 64 c 100
    0.436 61.1 c 99.13
    0.437 60.7 c 92.9
    };
\end{axis}
\end{tikzpicture}
\begin{tikzpicture}[trim axis left, trim axis right]
\begin{axis}[%
    title=STOP,
    title style={at={(1,1)},
                   anchor=north east,
                   draw=black,fill=white,
                   yshift=-6pt,},
    xmin=.395, xmax=.505,
    ymin=53.3, ymax=63,
    width=5.6cm, height=6.cm,
    grid=both,
    xtick distance=.02,
    ytick distance=2,
    xlabel=\textbf{LPIPS},
    ylabel style={xshift=-9pt, yshift=-15pt},
    xlabel style={yshift=8pt,},
    yticklabels={,,},
    tick label style = {font=\small},
    scatter, mark=*, mark size=1.5pt,
    scatter/classes={a={Mahogany}, b={MidnightBlue}, c={ForestGreen}},
    scatter src=explicit symbolic,
    nodes near coords style = {font=\scriptsize, inner xsep=-1pt},
    nodes near coords*={\Label},
    visualization depends on={value \thisrow{sr} \as \Label},
    ]
    \addplot[
    only marks,
    nodes near coords style = {text=Mahogany, anchor=north east},
    ] 
    table[meta=class, x index=0] {
    lpips fid class sr
    0.499 57.8 a 99.2
    0.468 56.9 a 97.7
    0.454 55.2 a 94.7
    0.439 55.2 a 87
    };
    \addplot[
    only marks,
    nodes near coords style = {text=MidnightBlue, anchor=south east},
    ] 
    table[meta=class, x index=0] {
    lpips fid class sr
    0.446 56 b 98.9
    0.438 55.6 b 98.1
    0.429 55.1 b 97.3
    0.425 54.9 b 95.5
    0.418 54.1 b 85.9
    };
    \addplot[
    only marks,
    nodes near coords style = {text=ForestGreen, anchor=north east},
    ] 
    table[meta=class, x index=0] {
    lpips fid class sr
    0.45 60.4 c 99.9
    0.443 62.8 c 99.46
    0.436 60.5 c 91.78
    };
\end{axis}
\end{tikzpicture}
\end{flushright}
\vspace{-0.6cm}
\caption{\textbf{FID vs.\ LPIPS, labeled with Success rate} for \acro{} and STEEX \cite{steex} explaining a decision model trained on the 4 labels of BDD-OIA.
Data points are obtained with different values of $\lambda_\text{dist}$, yielding different trade-offs.
\textcolor{ForestGreen}{STEEX} is directly comparable to \textcolor{MidnightBlue}{\acro{} \textit{style-only}} as they operate in similar conditions. See \autoref{sec:experiments:quality}.
\vspace{-0.2cm}
}
\label{fig:fid-lpips-success-rate}
\end{figure*}

Counterfactual explanations are defined as images that are similar to the query image but effectively change the decision of the target model.
To verify that our explanations fit the criteria, we evaluate them with the following metrics:
1) the FID \cite{heusel2017fid} which measures if they are realistic images,
and 2) a perceptual distance $\text{LPIPS}$ \cite{zhang2018lpips} that measures the change magnitude between them and the query image.
We also compute the success rate corresponding to the percentage of counterfactual examples that are indeed classified in the target class by the decision model.
We compute these metrics for different values of $\lambda_{\text{dist}}$ that lead to different trade-offs and plot the scores in \autoref{fig:fid-lpips-success-rate}.
As STEEX explanations are constrained to style features, we compare it to our \acro{} \emph{style-only} setting.
Firstly, in terms of FID, we can observe that both versions of \acro{} obtain lower FID than STEEX: \acro{} produces explanations of better visual quality, even though it does not use any annotations as opposed to STEEX, which needs semantic maps, and handles spatial explanations that are more challenging to generate.
Secondly, in terms of LPIPS, for style-based explanations \acro{} \emph{style-only} stays closer to the query than STEEX.
This can be explained as STEEX latent representation is semantic-based, while ours is instance-based.
Our method is therefore able to seek finer-grained modifications, while STEEX tends to change the style for an entire semantic class even when only one instance is important to the decision model.
As for \acro{}, because LPIPS disproportionately weights structural modifications, the wide range of attainable values indicates that many degrees of changes can be obtained depending on the choice of $\lambda_{\text{dist}}$.
Some settings tend to move objects a lot, while other settings attain values similar to `\emph{style-only}' edits.
Finally, the very high achievable Success Rates displayed by all three models confirm that they will propose explanations in almost any situation.

We display examples of OCTET’s counterfactual explanations in \autoref{fig:qualitative}.
OCTET finds sparse spatial and/or semantic modifications to objects that strongly influence the output decision.
For example, we can see with \acro{} that the distance with the car in front plays a role when deciding for `Forward' (1st row) or `Stop' (2nd row).
On the other hand, \acro{} \emph{style-only} shows that for those same decisions, the  brake lights also are taken into account.
For going `Left' class (3rd row), the absence of the continuous line seems crucial.
On the 3rd and 4th row, we can see that \acro{} mixes both style (brake lights) and spatial (proximity) explanations.
On the same tasks, STEEX changes the style of the whole class, and not just the necessary instance.

\begin{figure}
    \centering
    \includegraphics[width=\linewidth]{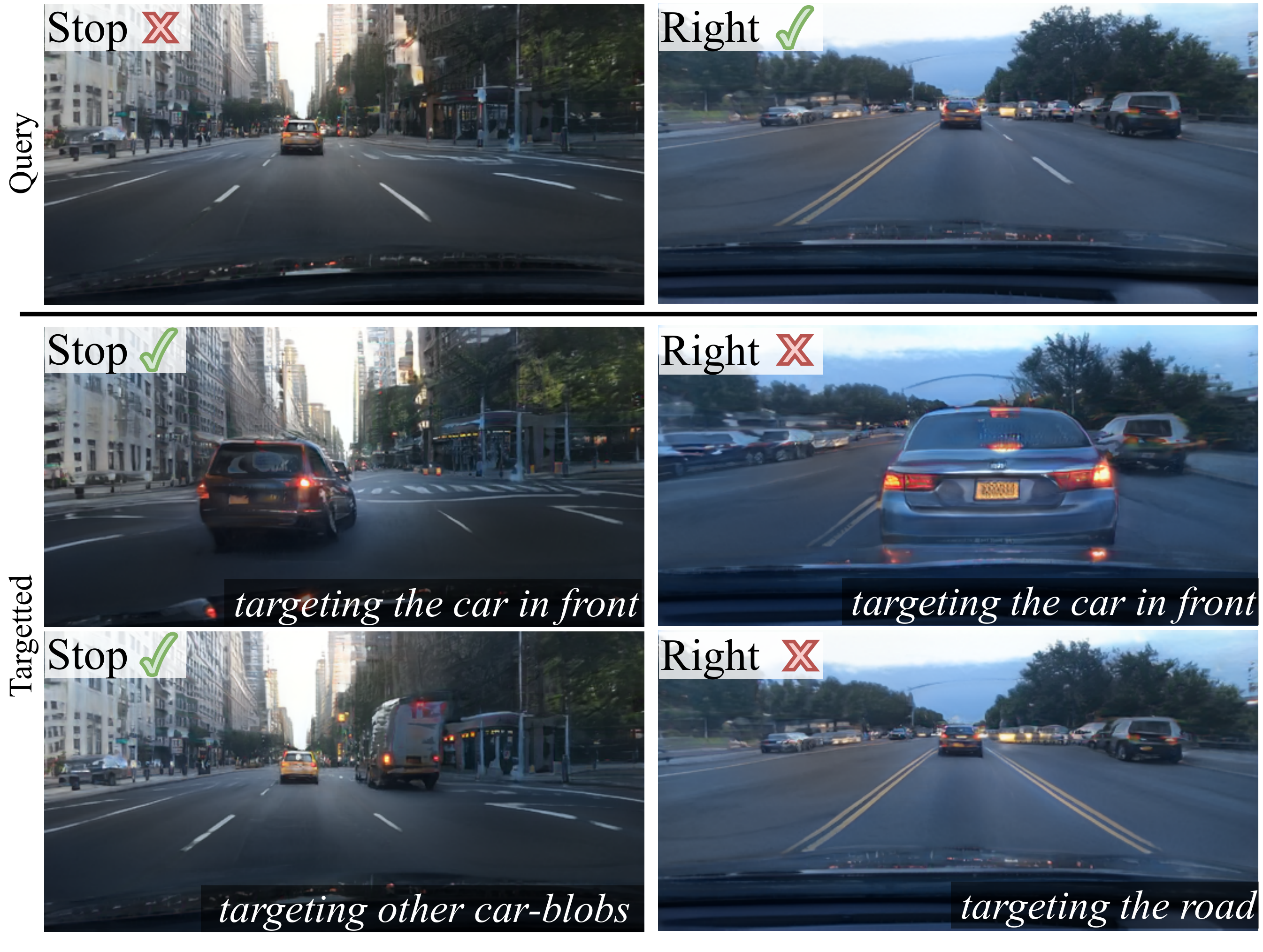}
    \caption{\textbf{Targeted counterfactual explanations}. 
    For each query image (top row), we target different blobs (2nd and 3rd row) to inspect the role  of corresponding objects.
    The output of the decision model is reported on the top right corner for each image.
    }
    \label{fig:targeted}
\end{figure}

\begin{figure*}
    \centering
    \includegraphics[width=\textwidth]{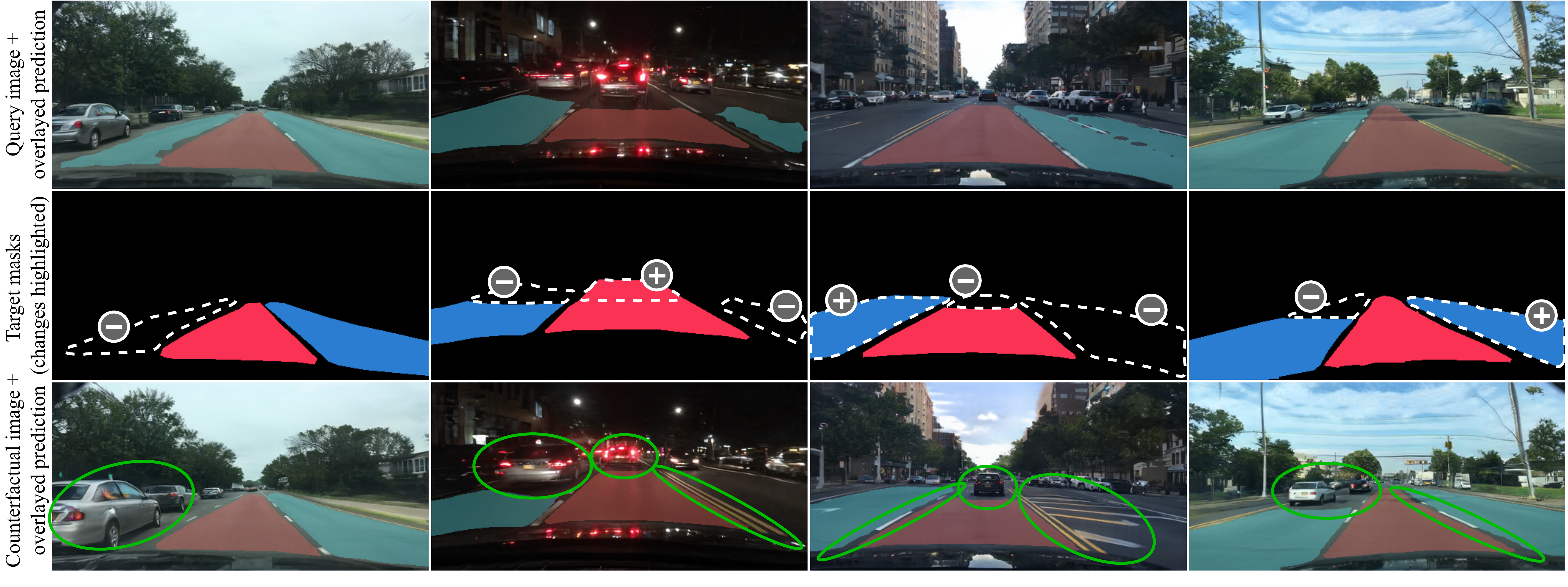}
    \caption{\textbf{Counterfactual explanations for a segmentation model.} 
    The segmentation model predicts the driveable area with primary and secondary lanes shown in red and blue respectively.
    The first row shows query images and segmentations by the model.
    In the second row, we add or remove parts of the original masks to build target masks (changes are shown with white dotted lines).
    The third row displays counterfactual explanations for the target masks and predictions on the new images.
    Green ellipses are manually added to highlight details.}
    \label{fig:semantic-segmentation}
\end{figure*}

\subsection{Targeted counterfactuals}
\label{sec:experiments:targeted}

Models that operate on compositional scenes can have their decisions influenced by many different factors, and we have shown that \acro{} can expose those factors in its counterfactual examples.
However, the user may want to test some particular hypothesis dealing with specific objects or areas of the scene.
Accordingly, the user can decide to optimize \autoref{eq:ldist-objects} only on selected blobs parameters.
Finding the blob corresponding to a specific object in the image is discussed in Appendix.
For instance, in the situation shown at the top left of \autoref{fig:targeted}, the car at the front seems to have its  brake lights on, but the decision is still not to `Stop'.
Targeting the front car (2nd row left image) reveals that the decision would have been `Stop' if the car was closer, and the brake light more clearly contrasted.
\acro{} can also target blobs that are currently inactive, corresponding to no visible object in the image.
In this example, doing so for car blobs shows that if another vehicle also had brake lights on, the model would also decide to `Stop' (3rd row left image).
This finding is unexpected, and could warrant further investigation of the decision model.
We show in \autoref{fig:targeted} and \autoref{fig:teaser} some other results we obtain by targeting specific blobs.

\subsection{Explaining a semantic segmentation model}
\label{sec:experiments:segmentation}

For complex tasks, like autonomous driving, decision models usually rely on multiple neural networks that address different subtasks (e.g., traffic sign recognition, depth estimation, motion forecasting).
In this context, it is important to explain models beyond classifiers \cite{vinogradova2020interpretable_segmentation,hoyer2019grid_saliency_semantic_segmentation}, a question not addressed in previous counterfactual methods.
Here, we show the versatility of our method as we extend it to explain a semantic segmentation network.
To do so, we use the same optimization problem described in \autoref{eq:cf-latent-objective} and \autoref{sec:method} but we change  $L_\text{decision}$ to an image segmentation loss.

In \autoref{fig:semantic-segmentation}, we show counterfactual explanations for a driveable area segmentation model.
More precisely, OCTET explains outputs of a DeepLabv3 \cite{deeplabv3} trained to segment images from \bdd{} into classes of driveable area: the primary lane, the secondary side lanes, and the background. 
We define the target masks by manually modifying predictions made on the query images as illustrated in the second row.
We see in the obtained counterfactuals, in the third row, changes in road markings and vehicles' positions.
They confirm that the segmentation model will consider as driveable a lane behind broken white lines (col.\,3) but not behind a double yellow line (col.\,2), especially if, in addition, the lane is hatched (col.\,3).
If there is a car on a driveable lane, the model can detect that the occupied portion is non-driveable while considering that the rest of the lane remains accessible.
With these examples, \acro{} provides explanations for both large semantic regions, e.g., when we add or remove an entire region in the target mask, or small subparts, e.g., when we slightly crop or extend a region.

\subsection{User-study to assess explanations}
\label{sec:experiments:user_study}

Metrics presented in \autoref{sec:experiments:quality} measure how well the produced examples match the definition of counterfactual explanations.
We want here to assess how \emph{useful} the explanations are for a user to better understand a decision model.
However, designing such an evaluation is challenging. 
Indeed, it has to avoid the many biases that appear \cite{nguyen2021effectiveness_attribution,doshi2017towards}.
Moreover, we usually do not have access to the inner workings of a model, and therefore we cannot compare users' insights on its behavior against a ground truth.
To overcome these challenges, we take inspiration from recent work on the usefulness of saliency maps as explanations \cite{fel2022userstudy} and we adapt its evaluation protocol to counterfactual explanations.

The goal is to measure if accessing counterfactual explanations on some examples helps users to better \emph{understand} the model.
Grounded within previous literature \cite{meaningful_perturbation,hase2020evaluating_xai,kim2016examples_not_enough}, we instantiate \emph{understand}
as `being able to predict the model's output for new instances', a concept known as \emph{simulatability}.
In particular, our user-study has two %
phases:
\begin{enumerate}[leftmargin=15pt,topsep=0pt,itemsep=0pt,parsep=0pt,partopsep=0pt]
    \item %
\textit{Observation phase}: participants are shown a series of triplets, each being composed of (1) a query image, (2) the model's decision on that image, and (3), a counterfactual explanation for the decision. %
\item \textit{Questionnaire phase}: participants are presented with new images and are asked to guess the output of the model on these images.
We stress that during the questionnaire, no counterfactual images are shown. 
\end{enumerate}

To measure the impact that counterfactual explanations of the first phase have on the users' ability to predict the model output on new samples (second phase), we use a control group. %
This control group, composed of different participants, undertakes the exact same tasks except that counterfactual explanations are \emph{not} shown during the first phase.
The decision model used in this study is a classifier trained on BDD-OIA \cite{bdd_oia} to predict the binary `Turn Right' label.
Besides, the decision model is trained, by design, with a flaw: the presence of a car on any side of the road can impact the decision, not only obstacles on the right side.
As discussed further below, this bias is introduced to study if users are able to identify model defects thanks to counterfactuals.
More details are provided in the appendix.

\begin{table}
    \centering
    \resizebox{\columnwidth}{!}{%
    \begin{tabular}{@{}l c c c@{}}
        \toprule
        & Cohort size & Replication & Bias Detection \\
        \midrule
        Group w/ \acro{} & 20 & \textbf{70\%} & \textbf{65\%} \\
        Group control & 20 & 52.5\% & 0\% \\
        \bottomrule
    \end{tabular}
    }
    \caption{\textbf{Results of the user-study.}
    The performance of users who had access to explanations (`w/ \acro{}') in the observation phase is compared to that of users who didn't (`control').
    `Replication' measures the accuracy of the users in predicting the model's output on new samples and `bias detection' is the proportion of users who found the spurious correlation in the model.
    \label{tab:user-study}
    }
\end{table}

\smallskip\noindent\textbf{Replication score.}
We first measure the `replication score': the proportion of correct match between the user's answer and the prediction of the decision model, averaged over participants of the group. 
We report these results in \autoref{tab:user-study} and observe that participants in the group accessing counterfactuals during the observation phase can successfully guess the decision model's prediction 70\% of the time while the control group's performance is 52.5\%, barely above random.
A statistical unpaired t-test, with as null hypothesis that the counterfactual explanations have no effect, confirms the significance of our result with a p-value of 0.0028.
This demonstrates the usefulness of \acro{} in understanding the decision model as participants with explanations could better anticipate the model's behavior on new instances.

\smallskip\noindent\textbf{Bias detection.}
Another promise of counterfactual explanations is to enable bias detection.
We then want to assess whether, by looking at explanations, users can tell if and how a decision model is biased.
In that direction, previous works dealing with human face datasets train biased models by artificially confounding attributes, e.g., smile and age, in the training set.
Then, they develop bias detection benchmarks that measure, by means of an oracle neural network, how well attributes remain confounded in the counterfactual explanations \cite{progressive-exaggeration,dive,dime}.
However, a serious limitation of that benchmark is that it keeps end-users out of the evaluation loop and it cannot tell whether or not a user will be able to uncover the spurious correlations by looking at the counterfactual explanation.
Moreover, detecting the issue with the model can be especially difficult when the bias is not obvious or its presence not suspected in the first place.
Instead of an automated evaluation, at the end of the questionnaire, we asked the participants to describe in free text what they had understood from the model.
The goal is to prompt them to mention the issue we embedded in the decision model, i.e., that the presence of a car on the left-hand side of the image factors in the `Turn Right' decision.
The `bias detection' score simply counts, for each group, the proportion of participants that used the word `left' in their answers.
In \autoref{tab:user-study}, we see that 65\% of the participants that had access to the explanations from OCTET identify the spurious correlation learned by the model, while none of the users that did not have the explanations (control group) mentions it.
This clearly shows that our method is useful to pinpoint issues with the decision model.
Also, the fact that not every participant detected it confirms our suspicions against automated bias detection benchmarks: the presence of the information in the counterfactual examples is not sufficient to ensure that the issue will be found in practice.

\section{Conclusion}
\label{sec:conclusion}

In this work, we presented \acro{}, a generative method to produce counterfactual explanations for deep vision models working on complex scenes.
Thanks to an object-centric representation of the scene,
\acro{} is able to process complex compositional scenes and to assess the impact of both the spatial positions and the style features of each object.
Our tool allows in-depth exploration of the different explaining factors by letting the user adjust on the fly the weights of the different search directions independently.
We evaluated our method thoroughly on real driving scenes from \bdd, with an improved evaluation benchmark compared to previous works, including a human-centered study that confirmed the usefulness of the approach. 

\section*{Acknowledgements}
This work was supported in part by the ANR grants VISA DEEP (ANR-20-CHIA-0022) and MultiTrans (ANR-21-CE23-0032).
Authors would like to thank the voluntary participants of the user-study, as well as Remi Cadene, Julien Colin, Thomas Fel, and Guillaume Jeanneret for helpful discussions.

{\small
\bibliographystyle{ieee_fullname}
\bibliography{biblio}
}

\appendix
\clearpage

\section{Technical Details}

\subsection{Decision models.}
The decision model used in the main paper is the same DenseNet \cite{densenet} as the one used in STEEX \cite{steex}. %
However, in addition to the `Move Forward' class, we also study `Stop', `Can turn Left', `Can turn Right' classes which were not discussed in STEEX.

\subsection{BlobGan backbone.}
The original BlobGAN \cite{blobgan} was trained on datasets of indoor scenes.
We present here the changes we made to train it on BDD.
Firstly, in order to generate rectangle images, we change the size of the input feature grid from $16 \times 16$ to $8 \times 16$, but increase the number of convolutional and upsampling blocks by 1, resulting in outputs images of resolution $256 \times 512$.
Secondly, the number of objects in the driving scenes is usually larger compared to the indoor scenes.
Therefore, we increase the number of blobs from $K=10$ to $K=40$.
However, multiplying the number of blobs by 4 increases the number of layout network parameters.
To keep the complexity reasonable, we decrease the size of the feature vectors describing the blobs from $d_\textit{in}=768, d_\textit{style}=512$ to $d_\textit{in}=d_\textit{style}=256$.
This model has 69.9M parameters in the generator (28.6M in the layout network and 41.2M in the synthesis network).
We trained the generator for 17 days (1.5M iteration) on one NVIDIA A100 GPU with a batch size of 10.

\subsection{Training the encoder}

The encoder $E$ is trained to predict the blob parameters from input images.
Its architecture is similar to that of the discriminator of StyleGAN-2, except for the output size of the last layer.
One challenge of predicting the blob parameters is that we have a large number of blobs, leading to a large number of trainable parameters in the encoder.
To avoid this issue, we opt to have the encoder not output directly the blob parameters but instead the features of the penultimate layer of the layout network.
The output layer of the encoder is then a vector, which we denote $h$, of size 1024;
$h$ can easily be mapped to the blob parameters by forwarding them through the last layer of the layout network.

The encoder is first pre-trained using generated images with the following reconstruction objective:
\begin{equation}
    \begin{split}
    L^\text{pretrain}_\text{encoder} & = L_{2}(h, E(G(h))),\\
    \end{split}
    \label{eq:supp:encoder_pretraining}
\end{equation}
where $h$ is obtained by sampling a noise vector and forwarding it through the first layers of the layout network.
For the sake of clarity, we omitted the layout network final layer that has to be applied to $h$ before being used as input to $G$.
Pre-training is done for 150k iterations using ADAM optimizer with a learning rate of 0.005 and batch size of 8.

Then, in the second stage, the encoder is fine-tuned on real and generated images with the following objective:
\begin{equation}
    \begin{split}
    L^\text{finetune}_\text{encoder} & = L_{2}(x, G(E(x)))\\
                    & + \lambda_\text{LPIPS} L_\text{LPIPS}(x, G(E(x)))\\
                    & + \lambda_\text{latent} L_{2}(h, E(G(h))\\
                    & + \lambda_\text{decision} L_{2}(f_M(x), f_M(G(E(x)))),\\
    \end{split}
    \label{eq:supp:encoder_training}
\end{equation}
where $h$ is obtained in the same way as during pre-training of the encoder, and $x$ by sampling from the dataset.
This objective focuses on three different aspects of image inversion.
First, we have to make sure that images that we encode as latent parameters with $E$ can be reconstructed by re-applying the generator using these latents.
This cycle consistency is enforced by a perceptual $L_\text{LPIPS}$ loss \cite{zhang2018lpips} and the $L_2$ loss between real and reconstructed images.
Second, we also ensure that generated images $G(z)$ can be encoded back into latent space using an $L_2$ loss between the generative latent parameters $z$ and their estimation by the encoder $E(G(z))$. This helps keeping the latents predicted by the encoder in the generator domain; this term is only used for generated images.
Finally, to encourage the reconstructed image to be faithful to the decision model, we use an $L_2$ distance between the features $f_M(x)$ of the input and the reconstructed image $f_M(G(E(x))$.
This term makes sure to preserve the features that are important to the decision model $M$.
Those features must contain both the decision taken by the decision model, but also capture the perceptual semantics that led to the decision.
In particular, for the DenseNet decision model that we have, we consider activations at the last convolutional layer of the decision model (DenseNet).
This is to ensure that the decisions are kept unchanged on the reconstruction but also that features that led to those decisions remain the same.
This finetuning stage is conducted with ADAM, for 150k steps, with a learning rate of 0.002.
Hyper-parameters are found after coarse manual inspection on some qualitative samples, $\lambda_\text{LPIPS} = 1$, $\lambda_\text{latent}=0.1$, $\lambda_\text{decision}=0.05$.
The choice of these hyper-parameters is not critical as obtained latent codes $z$ are then refined in an optimization phase, as explained in the main paper (\autoref{sec:method:image-inversion}).

\newcolumntype{P}[1]{>{\centering\arraybackslash}p{#1}}

\begin{table}
    \centering
    \resizebox{\columnwidth}{!}{%
    \begin{tabular}{@{}lP{1.5cm}P{1.5cm}P{1.7cm}@{}}
         \toprule
         & \phantom{ok} \newline FID ($\downarrow$) & \phantom{ok} \newline LPIPS ($\downarrow$) & Decision\newline preserv.\ ($\uparrow$) \\
         \midrule
         \autoref{eq:linversion} & 53.3 & 42.3 & 91.3\% \\
         \hspace{0.3cm} w/o\ $L_\text{LPIPS}$ (image) & 50.2 & 56.2 & 92.8\% \\
         \hspace{0.3cm} w/o\ $L_2$ (image) & 57.0 & 42.7 & 91.4\% \\
         \hspace{0.3cm} w/o\ $L_2$ (decision feat.) & 54.4 & 41.5 & 73.0\% \\
         \hspace{0.3cm} w/o\ $L_2$ (latent) & 52.0 & 41.8 & 91.3\% \\
         \bottomrule
    \end{tabular}
    }
    \caption{\label{tab:supp:ablation_reconstruction}\textbf{Ablation of the image inversion optimization process} (\autoref{eq:linversion}). The `Decision preserv.' is the percentage of images that yield the same decisions as their reconstruction using $M$.}
\end{table}

\subsection{Time complexity}
OCTET takes $\sim$28s per counterfactual image, in a batch of 16. The inversion step amounts to 85\% of that time and the CF optimization 15\%.

\section{Further ablations}
We present in this section ablation studies for the image inversion optimization process. In particular, we ablate the different terms of the loss.
Moreover, we recall that the initial values in this optimization process are given by the output of an encoder, and we ablate its training objectives as well.
In addition to the visual reconstruction metrics (FID, LPIPS and pixel-wise distances), we also evaluate the semantic fidelity of the reconstructions with a `Decision preservation' score. More precisely, the score measures the proportion of reconstructions that yield the same decision as the original image when presented to the model $M$.
This property is important as decision switches that occur during the reconstruction phase are guided neither by the target class nor by the studied model $M$ and therefore can hinder the downstream task of creating a reliable counterfactual explanation.

\subsection{Image inversion}

We present in \autoref{tab:supp:ablation_reconstruction} an ablation study of the terms of \autoref{eq:linversion} that we recall below:
\begin{equation}
\begin{split}
    \phi^q, \psi^q = \argmin\limits_{\phi,\psi} ~ & L_\text{LPIPS}(G(\phi, \psi), x^q)) \hspace{-0.07cm} + \hspace{-0.07cm} L_{2}(G(\phi, \psi), x^q) \\[-0.25cm]
    & \hspace{1cm} + L_2(f_M(x^q), f_M(G(\phi, \psi)))\\
    & \hspace{1cm} + L_{2}((\phi, \psi), E(x^q)).
\end{split}
\raisetag{0.5cm}
\label{eq:supp:linversion}
\end{equation}

We observe that the first two terms improve FID and LPIPS while the last two do not seem to influence those scores.
However, without the third term, we note a huge drop (91.3\% vs.\ 73.0\%) in the number of reconstructed images that conserve the model's original decision. The last term is a standard safeguard in the literature \cite{zhu2020indomainganinversion}: although there is no improvement in terms of LPIPS and FID, we %
observed that, without it, some objects were lost in the reconstruction, especially grey cars that blend in the road.

\subsection{Encoder training}

\begin{table}
    \centering
    \resizebox{\columnwidth}{!}{%
    \begin{tabular}{@{}lP{1.5cm}P{1.5cm}P{1.5cm}P{1.7cm}@{}}
         \toprule
         & \phantom{ok} \newline LPIPS ($\downarrow$) & \phantom{ok} \newline $L_2$ ($\downarrow$) & \phantom{ok} \newline $L_1$ ($\downarrow$) & Decision \newline preserv. ($\uparrow$) \\
         \midrule
         Pretrained enc. (after\,\autoref{eq:supp:encoder_pretraining}) & 57.2 & 0.230 & 0.324 & 63.1\% \\
         \midrule
         Finetuned enc. (after\,\autoref{eq:supp:encoder_training}) & 54.6 & 0.175 & 0.278 & 71.5\%\\
         \hspace{0.5cm} w/o\ $L_2$ on $f_M$ & 54.8 & 0.176 & 0.279 & 61.4\%\\
         \hspace{0.5cm} $L_1$ instead of $L_2$ & 55.3 & 0.186 & 0.285 & 66.3\% \\
         \bottomrule
    \end{tabular}
    }
    \caption{\label{tab:supp:ablation_encoder_training}\textbf{Ablation of the finetuning process of the encoder} (\autoref{eq:supp:encoder_training}). The `Decision preserv.' is the percentage of images that yield the same decisions as their reconstruction using $M$.}
\end{table}

\autoref{tab:supp:ablation_encoder_training} reports results for the ablation of the encoder training losses (\autoref{eq:supp:encoder_pretraining} and \autoref{eq:supp:encoder_training}).
The loss on features $f_M$ helps preserve the original decision of $M$ while keeping reconstruction quality.
Using it, 71.5\% of auto-encoded images yield the same decision as the input vs.\ 61.4\% without.
Also, we chose the $L_2$ pixel loss as in the BlobGAN paper. With $L_1$, results are slightly degraded.

\begin{table}
    \centering
    \begin{tabular}{@{}lcc@{}}
        \toprule
        & FID ($\downarrow$) & LPIPS ($\downarrow$) \\
        \midrule
        STEEX & 60.2 & 0.435 \\
        OCTET & \textbf{53.3} & \textbf{0.423} \\
        \bottomrule
    \end{tabular}
    \caption{
    \label{tab:sup:inversion}
    \textbf{Reconstruction capacities of OCTET and STEEX.}
    We check the quality of reconstruction with FID (is the reconstructed image realistic?) and LPIPS (is the reconstructed image close to the input image?). %
    The values below were computed on the validation set of BDD segmentation dataset (1000 images).
    Note that STEEX uses ground truth segmentation masks as additional input.
    }
\end{table}

\section{Reconstruction quality}

We compare the ability of OCTET and STEEX to reconstruct real images. %
To measure the reconstruction quality, we use the FID \cite{heusel2017fid} between all reconstructions and the set of real query images.
We also measure the mean $L_\text{LPIPS}$ \cite{zhang2018lpips} distances between all pairs of real and reconstructed images.

In \autoref{tab:sup:inversion}, we report FID and LPIPS scores for reconstructed images.
Overall, we observe that the backbone used in OCTET and our inversion strategy (encoder + optimization) leads to images that are closer to the input image (lower LPIPs) but also more realistic (lower FID) which means that they are closer to the original data distribution. %
Moreover, we stress that OCTET does not use any ground-truth segmentation maps, unlike STEEX.

\begin{figure*}
    \centering
    \includegraphics[width=\textwidth]{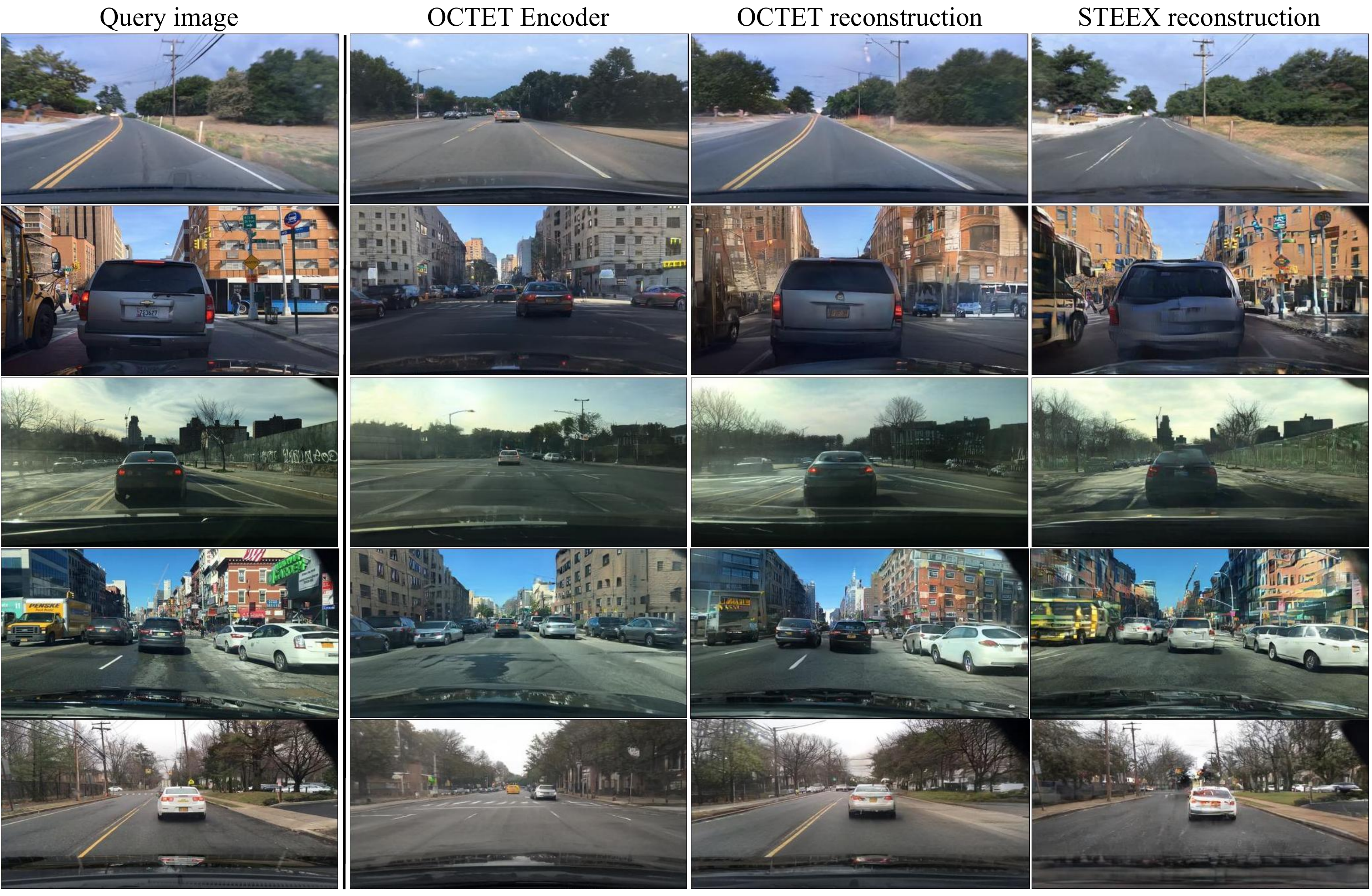}
    \caption{\textbf{Examples of inversion results for OCTET (ours) and STEEX \cite{steex}.} }
    \label{fig:supp:inversion}
\end{figure*}

In \autoref{fig:supp:inversion}, for some query images (1st column), we show some qualitative results of:
\begin{itemize}
    \item images obtained with the encoder $E$ which are then used as starting points of the optimization described in \autoref{eq:linversion} in the main paper (2nd column); 
    \item OCTET reconstructions, which are images obtained after further optimization with \autoref{eq:linversion} (3rd column);
    \item STEEX reconstructions (4th column).
\end{itemize}

Overall, while the encoding step is able to grasp the rough structure and colors of the scene, we can observe that the optimization steps that follow are key to get the fine and precise details of the scene. 
Moreover, we can see that STEEX reconstructions sometimes miss important details (lines and brake lights) that are well captured in the latent space we use for OCTET. 
Finally, on some small background details, STEEX reconstructions are sometimes more accurate and faithful to the query image (e.g., the shape of trees).
However, STEEX makes use of semantic maps that indicate the semantic class of each pixel. These maps are strong guides but are very costly to produce for end-users. 

\subsection{Discussion on sparsity of changes}
\begin{figure*}
    \centering
    \includegraphics[width=0.8\textwidth]{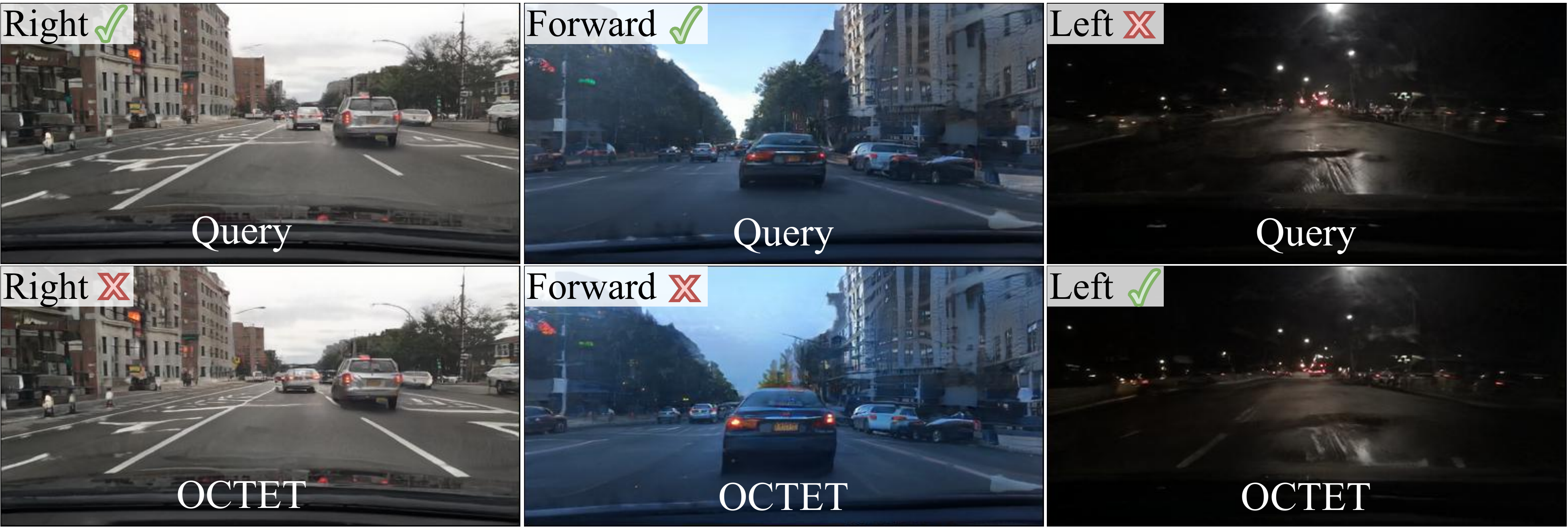}
    \caption{\label{fig:sup:non_cherry_picked_cf}
    \textbf{Counterfactual explanations on generated query images.} Neither the images nor the target classes are cherry-picked.
    }
\end{figure*}

Sometimes, counterfactual explanations do not display sparse changes with respect to the original query image.
In fact, the main source of non-sparsity comes from the challenges of GAN inversion.
Despite our efforts and contributions allowing us to outperform previous works in this regard, our reconstruction already introduces changes (see \autoref{fig:supp:inversion}).
The counterfactual optimization step itself does not degrade LPIPS further (see \autoref{fig:fid-lpips-success-rate} in the main paper and \autoref{tab:sup:inversion}).
In \autoref{fig:sup:non_cherry_picked_cf} (not cherry-picked), we illustrate the sparsity of the CF optimization itself by starting from generated queries to build the CF, thus not necessitating a reconstruction step.
As the sparsity in this step is satisfactory, we decided not to add a pixel-level regularization that would also heavily penalize explanations involving object displacement.

\section{Finding the blob index to target}
\label{sec:experiments:blobgan}

We discussed in the main paper how to assess the importance of specific blobs with OCTET.
But for practical use, we need to identify which blob corresponds to a given object in the scene.
We here explain how, taking advantage of BlobGAN \cite{blobgan}, the generative backbone we use, we are able to find the blob identity.

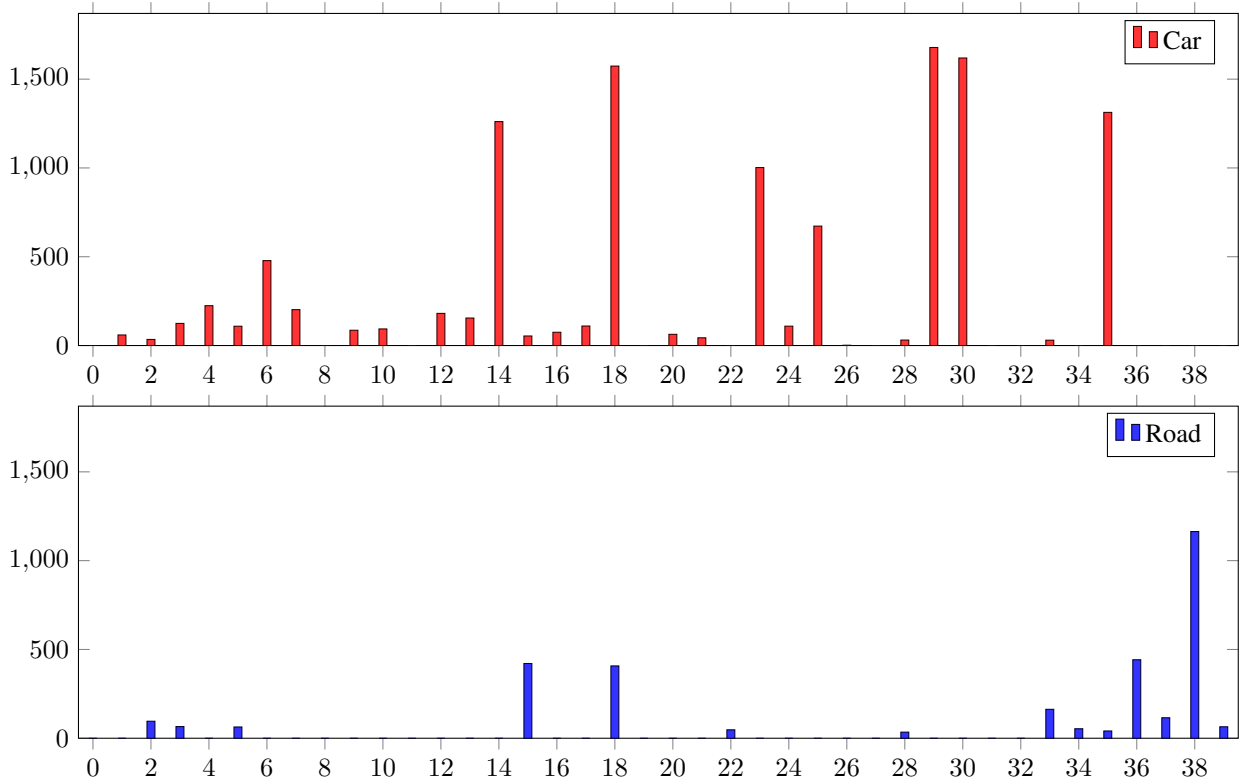
\begin{figure*}
    \centering
    \scalebox{1}{\begin{tikzpicture}
 
\begin{axis} [ybar,height=6cm,
    width=17cm,
    bar width = 3pt,
    xmin = 0, 
    xmax = 39, 
    ymin=0,
    ymax=1700,
    enlarge y limits = {value = .1, upper},
    enlarge x limits = {abs = .5},
]

\addplot[red!20!black,fill=red!80!white] coordinates {(0, 0.0)
(1, 59.47)
(2, 34.325)
(3, 124.555)
(4, 224.385)
(5, 108.715)
(6, 478.225)
(7, 202.085)
(8, 0.0)
(9, 86.02)
(10, 93.465)
(11, 0.0)
(12, 180.995)
(13, 154.91)
(14, 1261.015)
(15, 53.765)
(16, 74.72)
(17, 109.91)
(18, 1573.61)
(19, 0.0)
(20, 63.135)
(21, 43.65)
(22, 0.0)
(23, 1002.225)
(24, 109.04)
(25, 672.48)
(26, 1.78)
(27, 0.0)
(28, 30.785)
(29, 1678.27)
(30, 1619.13)
(31, 0.0)
(32, 0.0)
(33, 30.11)
(34, 0.0)
(35, 1312.875)
(36, 0.0)
(37, 0.0)
(38, 0.0)
(39, 0.0)};

\legend {Car};
 
\end{axis}
 
\end{tikzpicture}}
    \scalebox{1}{\begin{tikzpicture}
 
\begin{axis} [ybar,height=6cm,
    width=17cm,
    bar width = 3pt,
    xmin = 0, 
    xmax = 39, 
    ymin=0,
    ymax=1700,
    enlarge y limits = {value = .1, upper},
    enlarge x limits = {abs = .5},
]
 
\addplot[blue!20!black,fill=blue!80!white] coordinates {(0, 0.0)
(1, 0.0)
(2, 95.79)
(3, 65.695)
(4, 0.0)
(5, 63.4)
(6, 0.0)
(7, 0.0)
(8, 0.0)
(9, 0.0)
(10, 0.0)
(11, 0.0)
(12, 0.0)
(13, 0.0)
(14, 0.0)
(15, 420.785)
(16, 0.0)
(17, 0.0)
(18, 407.195)
(19, 0.0)
(20, 0.0)
(21, 0.0)
(22, 47.45)
(23, 0.0)
(24, 0.0)
(25, 0.0)
(26, 0.0)
(27, 0.0)
(28, 34.32)
(29, 0.0)
(30, 0.0)
(31, 0.0)
(32, 0.0)
(33, 162.89)
(34, 53.43)
(35, 40.87)
(36, 441.925)
(37, 115.375)
(38, 1163.96)
(39, 64.52)};

\legend {Road};
 
\end{axis}
 
\end{tikzpicture}}
    \caption{\textbf{Blobs semantic meaning.} We visualize the correlation between the blobs and the semantic classes 'car' (\textit{top}) and 'road' (\textit{bottom}). The x-axis displays blob ids, while values in the y-axis represent the mean number of pixels that are no longer from class $c$ when removing blob $k$ computed by sampling 200 different latent codes. The number of pixels for class $c$ is estimated using a pre-trained semantic segmentation network. }
    \label{fig:supp:blob-semantic}
\end{figure*}

\begin{figure*}
    \centering
    \includegraphics[width=\textwidth]{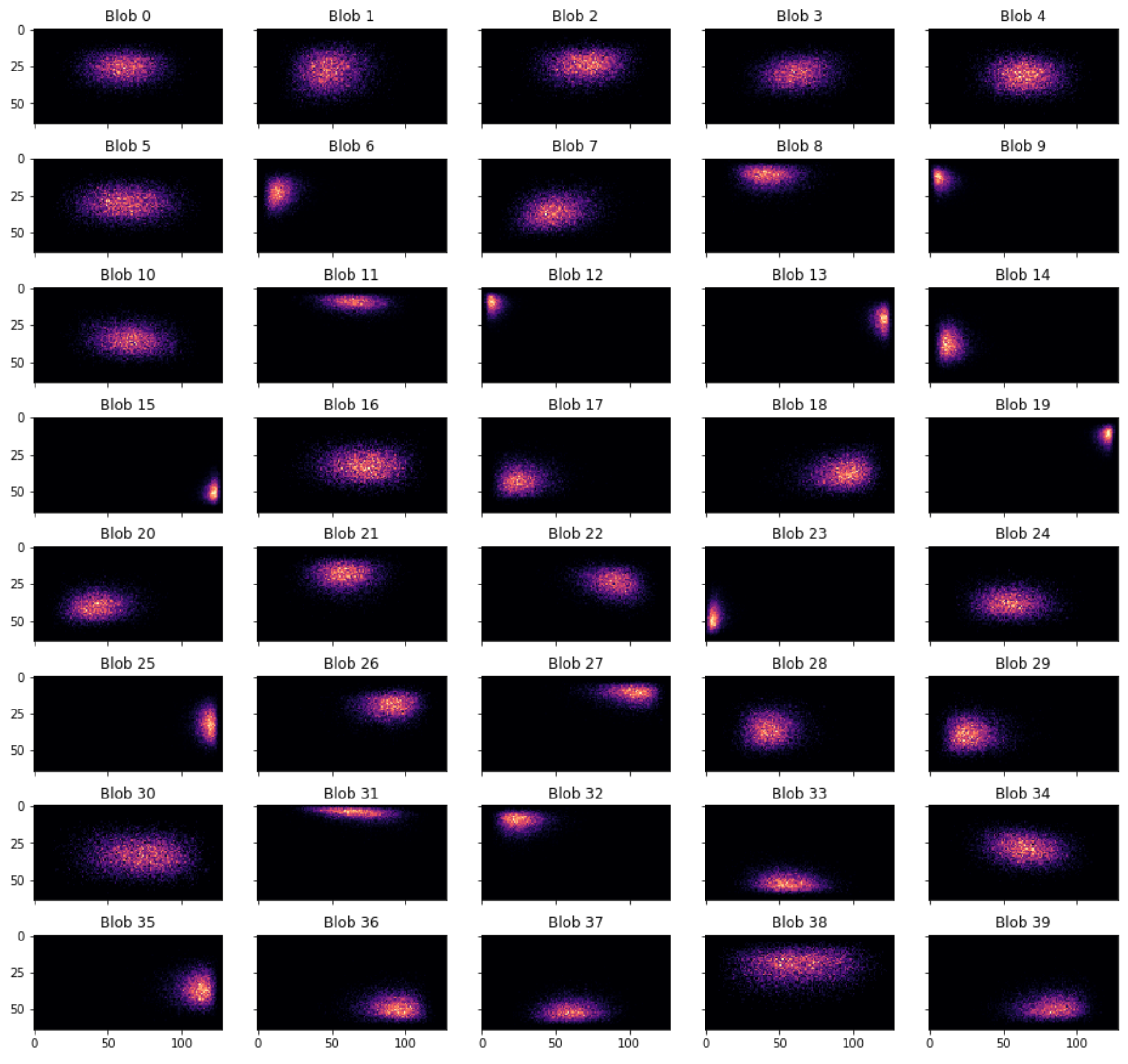}
    \caption{\textbf{Blobs spatial distribution.} We visualize the spatial distribution of blob centroids. By combining the information about the semantic meaning of the blobs and their spatial localization, we can precisely label main blobs (e.g., blobs representing the front car vs.\ blobs representing cars on the right).} 
    \label{fig:supp:spatial-distrib}
\end{figure*}

BlobGAN learns to represent scenes as a collection of blobs distributed on a canvas in an unsupervised fashion.
As discussed in the original paper for \emph{indoor scenes}, even without supervision from object positions and classes, the model learns to associate certain blobs with certain objects.
We observed that similar properties emerge when training the model on \emph{outdoor driving scenes}.
For instance, in \autoref{fig:supp:blob-semantic}, we visualize the correlation between blobs and the semantic classes for cars and roads.
We use a pre-trained semantic segmentation network to measure the number of pixels that disappeared from each class when removing a blob by setting its size to a negative number.
Using such figures, we are able to determine for instance that blobs 14, 18, 23, 29, 30, and 35 are very likely to correspond to car-blobs.
We then visualize in \autoref{fig:supp:spatial-distrib} the distribution of the spatial positions of the center of the blobs on the canvas.
To be clear, for each blob index, we plotted the location of its center and accumulated the plots over 10k generated images.
This figure shows that the blobs have a localized position which is consistent with the fact that the blobs have a semantic meaning.
Combining the location and the semantic class of the blob, we can infer that blob 30 is likely to correspond to a car in the middle of the image, while blob 35 is a car on the right for instance.

\begin{figure*}
    \centering
    \includegraphics[width=\textwidth]{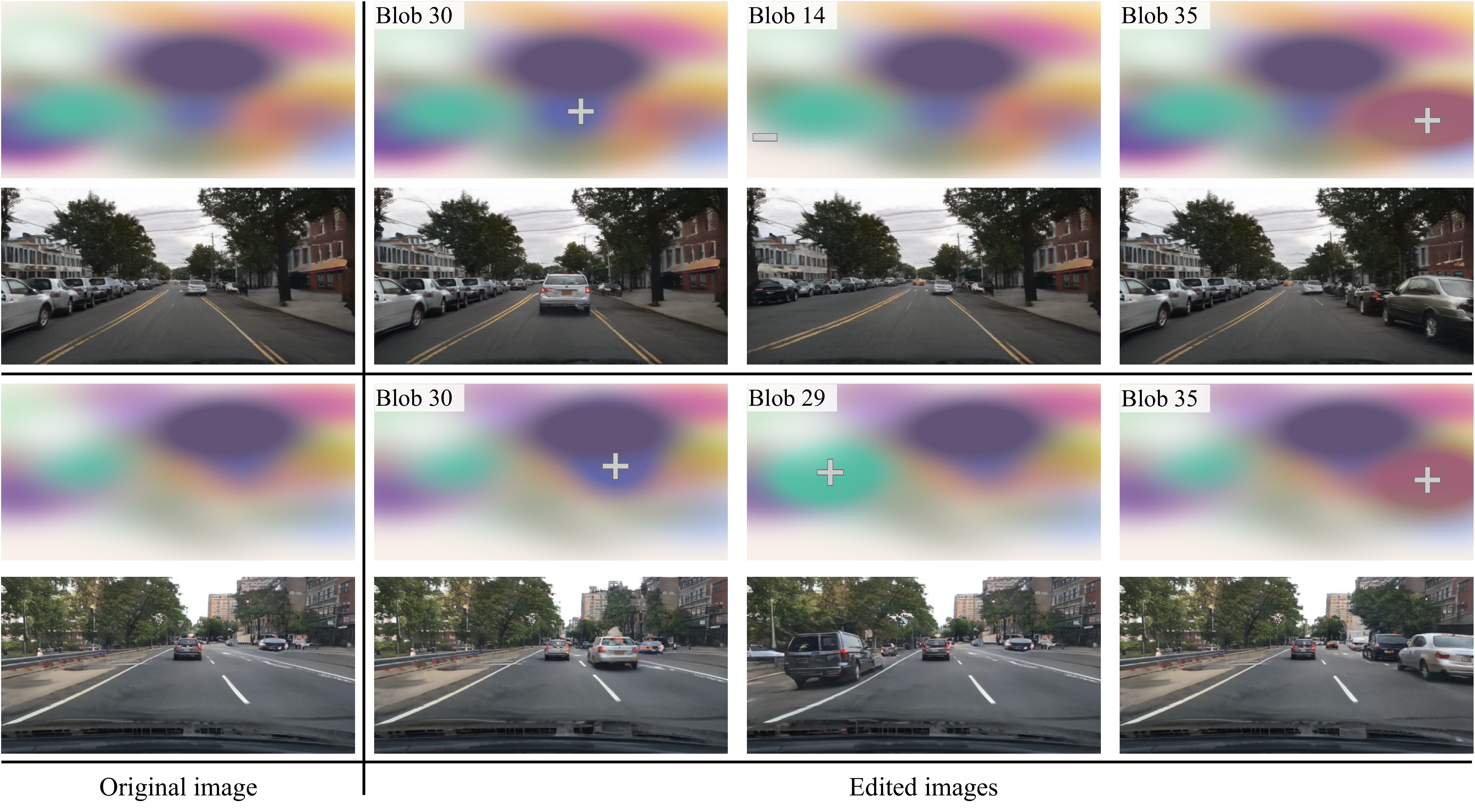}
    \caption{\textbf{Resizing cars.} We change the size of certain blobs representing cars. In addition to being able to change their size, we can make them appear and disappear. We stress that we are doing \emph{manual edition} by intervening on the size parameter of the blobs. This contrasts with other figures of the main paper where we are doing \emph{counterfactual explanations} and changes are automatically found with the optimization process to explain the decision of a model.
    }
    \label{fig:supp:edits-size}
\end{figure*}

\begin{figure*}
    \centering
    \includegraphics[width=\textwidth]{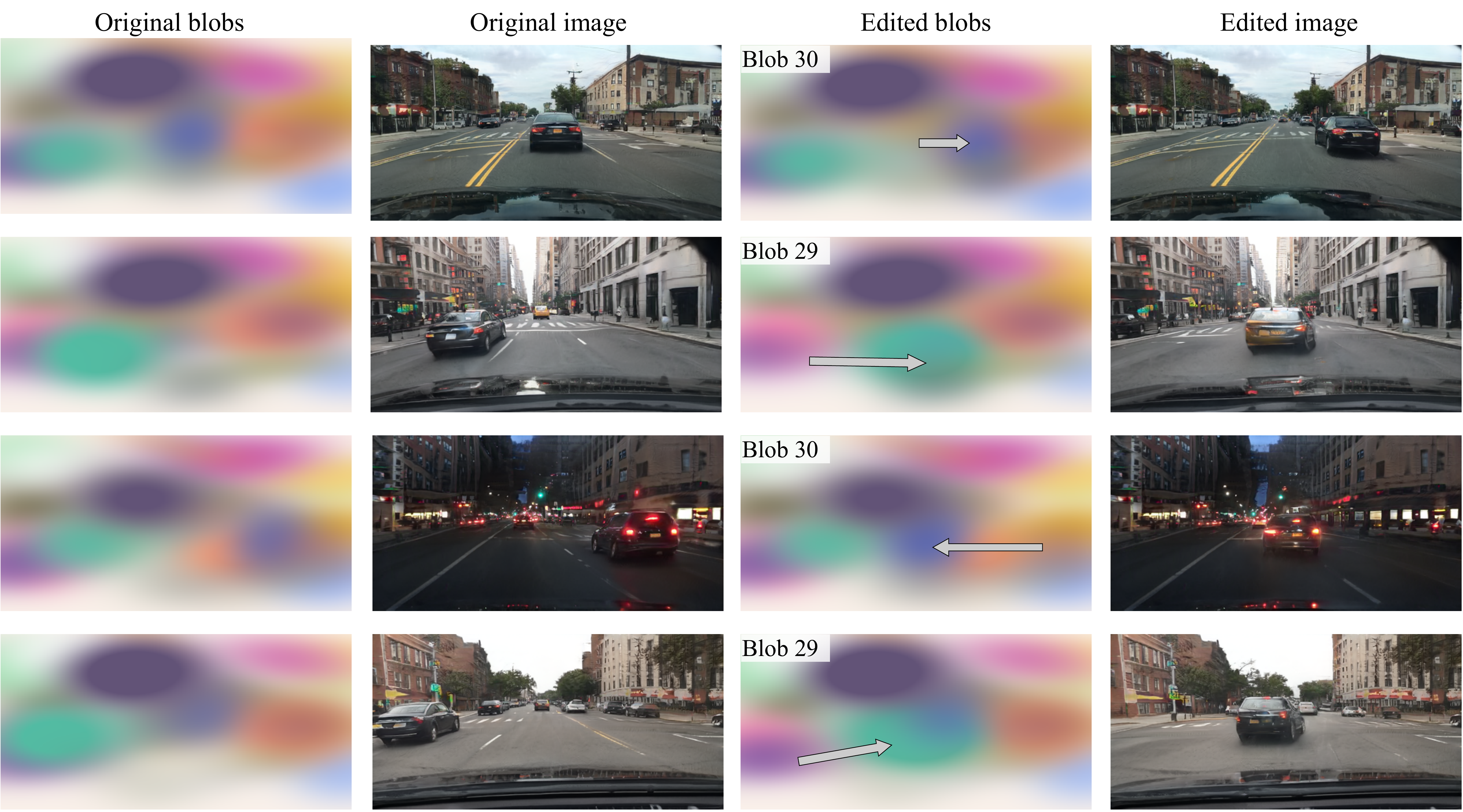}
    \caption{\textbf{Moving cars.} We change the position of certain blobs representing cars as pointed by the arrow by changing their centroid coordinates.
    We stress that we are doing \emph{manual edition} by intervening on the position parameters of the blobs. This contrasts with other figures of the main paper where we are doing \emph{counterfactual explanations} and changes are automatically found with the optimization process to explain the decision of a model.
    }
    \label{fig:supp:edits-position}
\end{figure*}

Using this knowledge, we can then directly intervene on the spatial parameters of the blobs to see how it affects the image and confirm our hypothesis.
We show the results of such manipulation in \autoref{fig:supp:edits-size} and \autoref{fig:supp:edits-position}.
These findings are consistent across images and make it fairly straightforward to identify the correct blob for object-targeted counterfactuals presented in the main paper.

\section{Details on the User Study}

\begin{figure*}
    \centering
    \includegraphics[width=\linewidth]{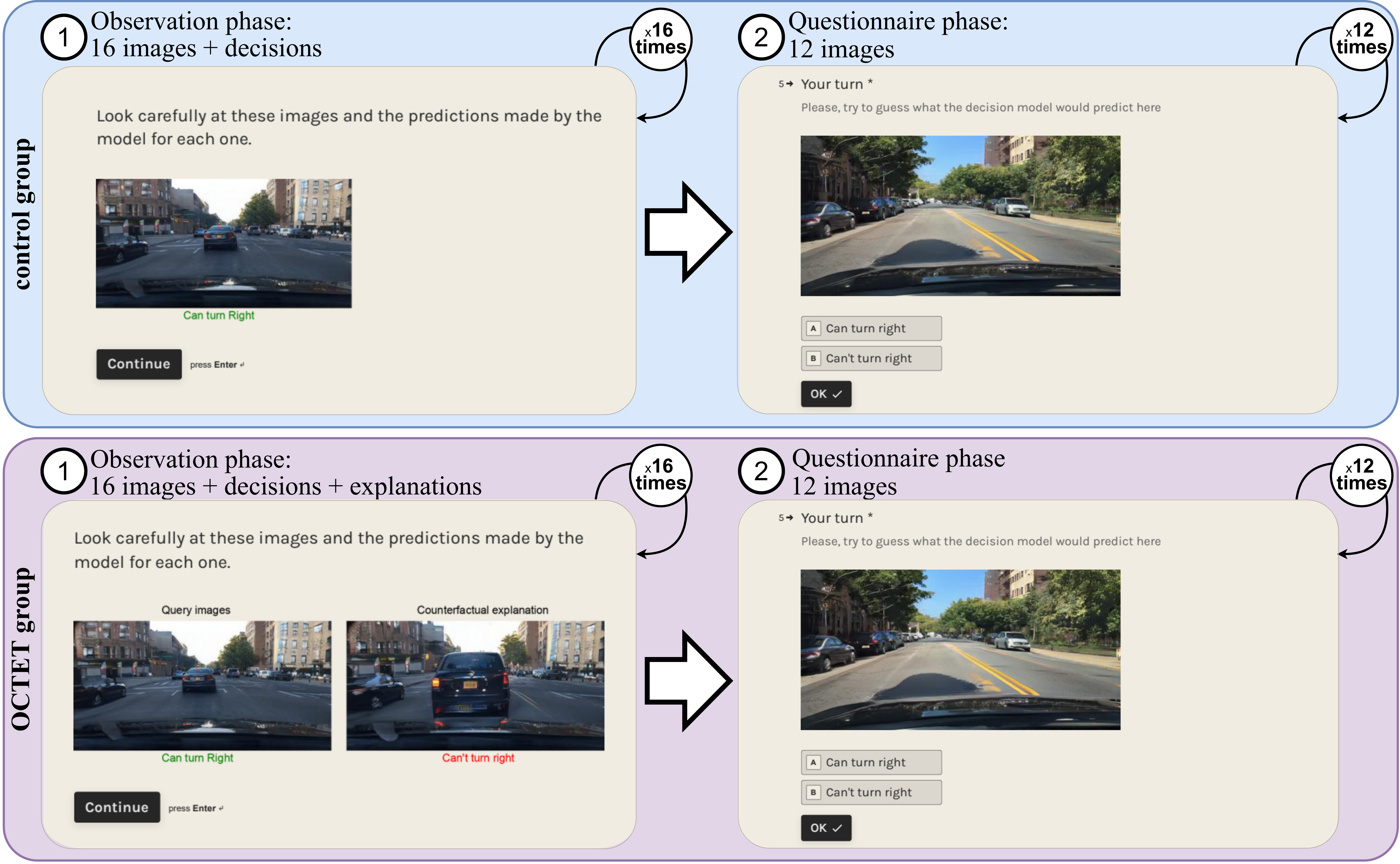}
    \caption{\textbf{User study overview.} The study is conducted as an online form in two phases. First, in the observation phase, the participant is shown examples to analyze the prediction of the model. Then, in the questionnaire, they are asked to guess the prediction of the model. The control group (\textit{bottom part}) is only shown images and associated decision in the observation phase. The OCTET group (\textit{top part}) has, in addition, access to counterfactual explanations.}
    \label{fig:user-study-forms}
\end{figure*}

\subsection{Protocol details}
Here, we describe more precisely the details of the user study (Sec.\ 4.5).
The experiment was conducted as an online form, with no interaction with any operator. The respondents are all voluntary, and we targeted participants that have familiarity with deep learning.
Participants are randomly split across the two groups (Control group and group with explanations).
We show in \autoref{fig:user-study-forms} a description of the study and the templates we used for the forms.

To build the online form, we needed 28 distinct query images: 16 were used for the observation phase and 12 for the questionnaire.
Those were randomly sampled, with no overlap between the two sets.
Unknown to the participants, we made sure that inside each set there are an equal number of true positives, true negatives, false positives, and false negatives.
This prevents the user's from over-relying on their prior knowledge of the task, and should also facilitate the detection of interesting behaviors of the model. 
For the observation images, we have to compute the predictions from the decision model and the counterfactual explanations.
The form for the test group contains all three elements while the form for the control group only contains the images and the decision, but no explanation.
For the questionnaire, we simply provide the raw images to both groups, without any additional information.
Using those same image sets for the two groups ensures that the comparison between them is more reliable.
However, it does not control for the variance induced by image selection.
To obtain more robust results, we build 5 versions for each form by randomly sampling 5 times the 28 query images.
This leads to a total of 10 variations of the forms (5 for the control group and corresponding 5 for the test group).
We randomly sample the assignment of one version of the form to each participant.

For the bias detection study, we first presented them with one additional test image, the same to all participants, for which they had to predict the output of the decision model.
We were not interested in their prediction this time, but we then asked them: "On the last image, please explain the factors that drove your choice in a few words".
We also asked them two additional questions:
\begin{itemize}
    \item "Did you manage to identify any particular behavior of the decision model?"
    \item "Did you manage to identify any problem or unexpected behavior in the decision model?"
\end{itemize}
The goal of those questions was to lead them to describe any peculiar behavior they would have inferred from the observation phase.

\subsection{Collected responses}
The collected answers to the free-form questions are presented in \autoref{tab:supp:responses}. Unknown to the participants, the decision model looked at the presence of cars on both the left and the right side of the road to make its prediction on the `Turn Right' label. No participant in the control group mentioned this issue while 65\% of those that had access to the OCTET explanations did.
\begin{table*}
    \centering
    \tiny
    \rowcolors{2}{white}{gray!25}
    \begin{tabularx}{\textwidth}{l X X X}
    \toprule
        Group
         & Question 1: On the last image, please explain the factors that drove your choice in a few words 
         & Question 2: Did you manage to identify any particular behavior of the decision model?
         & Question 3: Did you manage to identify any problem or unexpected behavior in the decision model? \\
    \hline
         Control 
         & white lanes visible right. A Car is also at the right of the screen but not a dense lane of cars
         & yellow lanes means can't turn right. If there is a dense lane of cars can't turn but if there is a few can turn. If the ground is covered with some white marks can turn.
         & yes the model does not understand if the right lane is occupied or going in the other direction. When the lane is free but there is a dense lane of cars parked the algorithm predicts that we can't turn right \\
         & Car detected on the right
         & The decision is "Can't turn right" when there are yellow ligns, or there is a car or an obstacle on the right.
         & The model does not turn right even when the road seems empty, or does turn right when the right path is very narrow \\
         & Presence of the car
         & The model focus mostly on ground lines and the presence of a car
         & Sometimes the model get confused when ground lines are not typical \\
         & 12th example looked the same, the front car being too close to the ego-car.
         & No
         & Similar paterns (e.g. yellow lines) didn't seem to be taken into consideration by the model \\
         & Black car on the right
         & Don't turn right if there is a car ahead on the right lane.
         & Not sure, but might not turn right if the car ahead (in the same line) is close.) \\
         & 
         & only a car in the same lane but the right lane is free --> "can't turn". Cars on both lane --> "can turn".
         & problem with yellow lanes \\
         & white line on the right
         & white line on the right
         & full white line on the right \\
         & yellow line, visibility of other cars, lighting/contrast
         & 
         & lighting affected the model a lot \\
         & The is a car on the right
         & If it sees a car on the right far away, doesn't turn right. If it's very close, doesn't. If it's a few meters ahead, checks that is not parked ?(brake lights on?) Or that the light is green? Not that much sensitive to double yellow lines
         & Bit very sensitive to double yellow lines \\
         & Right lane is clearly occupied by a car, the classifier can handle such situation
         & Hardly, maybe tend to say 'can't turn right' while right lane is open because he sees a car on a third lane (to the right of the lane to the right)
         & see previous answer \\
         & It seems that if the front car is too close, the prediction will always be can't turn
         & If there was a double Yellow line on the right, the model would always return false	
         & \\
         & It is a situation never seen before (car stop behind another, next to another), in doubt, you can't turn right.
         & Not really, it was mostly "right", but I didn't understand the pattern of the wrong predictions.
         & Yes, some predictions were wrong sometimes (e.g., crossing a full line). \\
         & due to the car on the lane next to our car
         & yes, I think the model is very well in detecting the two yellow parallel lines on the street indicating a prohibition of reel swapping
         & Yes, problem turn right is NOT turn 90 degree angle right but just turn right to the next lane as in reel swapping. I think this decision should be differentiated cause I find "turn right" confusing as it can mean both = also a problem for the model \\
         & 
         & no
         & no \\
         & Lines on road (solid vs dashed) + new objects on the right (Can't turn) + there is no car or a bit far (Can turn)
         & Not clear, It seems that if there is solid line on the right, a close car or new objects it can't turn
         & yes, many false negative \\
         & the road, the cars, the lines on the road (continuous/dashed..)
         & 
         & sometimes turning right is permitted by the model while there is a continuous line on the road \\
         & obstructing car on the right
         & free road, available turn further down the road = right turn ok; car present on the right lane = right turn not ok
         & the model does not understand the direction of traffic \\
         & The image is unlike the others. And in case of doubt, I assume that the model is conservative
         & Double lines make it not turn right
         & Does not pay too much attention to other cars \\
         & car too close
         & no
         & no \\
         & the car is too close from the other car
         & i think : car in front of the car and yellow line	
         & \\
        \hline
         OCTET
         & No car on the \textbf{left}
         & If a car is on the \textbf{left} the model predict can’t turn right
         & It seems that the model is disturbed by the car on the \textbf{left} \\
         & 
         & looking for cars on the \textbf{left}, or less often in front of our car
         & Model should look also look to the right. Often, a car on the \textbf{left} is not an issue to turn right \\
         & No cars stationed on the \textbf{left}, no continuous white band
         & The model might be disproportionately affected by lane width and parked cars
         & The model sometimes focus on irrelevant factors \\
         & Car in front and on the right too close to me
         & To decide if the car can turn right, the model is looking if there are closed cars on the right side, in front AND on the \textbf{left} side
         & Yes, to decide if the model can turn RIGHT, the model is also looking whether there is a car near on the \textbf{LEFT} side \\
         & Multiple cars, multiple brakes lights, car too close
         & Having a car on the \textbf{left} or multiple brakes is more likely to be "can't turn" prediction
         & Sometimes it says you can turn right although there does not seem to be a road on the right \\
         & Cars are too close and this seems to be a no no for the decision model
         & If there are cars too close it seems not to allow to turn right
         & \\
         & The car in front of me in close and the lights are on which seems to be understood by the model as "can't turn right"
         & When there is a car in the bottom \textbf{left} part of the image, it's tagged "can't turn right"
         & The results are very counter-intuitive because one would expect that 'can't turn right" should be the answer if there is a car or an obstacle on the right (not on the \textbf{left}) \\
         & 
         & 
         & \\
         & because the right lane is occupied
         & The presence of a double yellow line on the right prohibits turning right
         & I did not understand many examples where there was no double line and where the right lane seemed clear and where, however, turning right was prohibited. \\
         & Can't turn right when front vehicle is close and when there is a car on the \textbf{left} or on the right
         & Influcene the decision : Yellow line, distance with the front car, position of vehicles on the right or on the \textbf{left}.
         & The model does not consider the lateral space on the right side. \\
         & The car in front is too close, and there is a car on the right
         & To be able to overtake, we must not have cars overtaking us, nor cars that are too close, if possible good general visibility, in particular lines.
         & No \\
         & There is a car too the right which is close.
         & It does not want to turn when there is a car to the \textbf{left}
         & Yes, as said before \\
         & There is no car on the \textbf{left}, no double yellow can be seen
         & if the model detects an object car-shaped on the \textbf{left}, a double yellow line on the ground, or a long and large object to the ground on the right it doesn't turn right
         & it's decisions do not seem correct regarding driving ability \\
         & Car ahead too close
         & - double yellow lines: no turn - cars on the \textbf{left}: no turn - car ahead too close: no turn - need a car ahead to evaluate if there is the space to turn right
         & cars on the \textbf{left} line: no tun - need a car ahead to evaluate if the road is large enought to turn \\
         & car visible on the right
         & Yes when cars are closed either \textbf{left} or right the model predicts "can't turn right"
         & Yes, the understanding of the geographic position of other cars \\
         & car on the right, too close to my car
         & car on \textbf{left}/right close to my car. continuous white line on the right. Yellow line often on the right and sometimes on the \textbf{left}
         & can't turn on right when cars or lines on the \textbf{left} \\
         & In some images the model did not detect the car at the right hand part. SO here it seems "too far" to be detected by the model.
         & Seems to respect more the lines than the vehicles or pedestrians
         & Should prevent fro turn right when there is a car a the right... \\
         & Car on the right too close to the driver's car
         & Look for the presence of white line + presence of a car within a radius
         & Sometimes a car is too close on the right side and the model still consider it can turn right. Sometimes the prediction is just incorrect (yellow line and still predicts it can turn right) \\
         & There is a car next on the right that is too close to the initial car wanting to make a turn, so the model should detect that it is close enough and it's not safe to do a right turn .
         & Something i noticed is that when the line is continuous or not,
         & Yes the fact that is doesn't distinguish the nature of the line drawn in the ground \\
         & 1) there's a very close to the right. 2) the car just in front is very close.
         & failure to distinguish between right and \textbf{left}. When there's a closer car to the \textbf{left}, the decision is "can't turn right" even though the car can indeed turn right. (Naive question : are you using random flipping as data augmentation technique without changing the label during your training?)
         & failure to distinguish between right and \textbf{left}. When there's a closer car to the \textbf{left}, the decision is "can't turn right" even though the car can indeed turn right. (Naive question : are you using random flipping as data augmentation technique without changing the label during your training?) \\
         \bottomrule
    \end{tabularx}
    \caption{\textbf{Responses to the free-form questions in the user-study.} The participant had the option to leave the fields blank; they are still shown in the table. We highlight in bold every occurrence of the word `left'.}
    \label{tab:supp:responses}
\end{table*}

\section{Preliminary experiments on LSUN dataset}

\begin{figure*}
    \centering
    \includegraphics[width=0.8\textwidth]{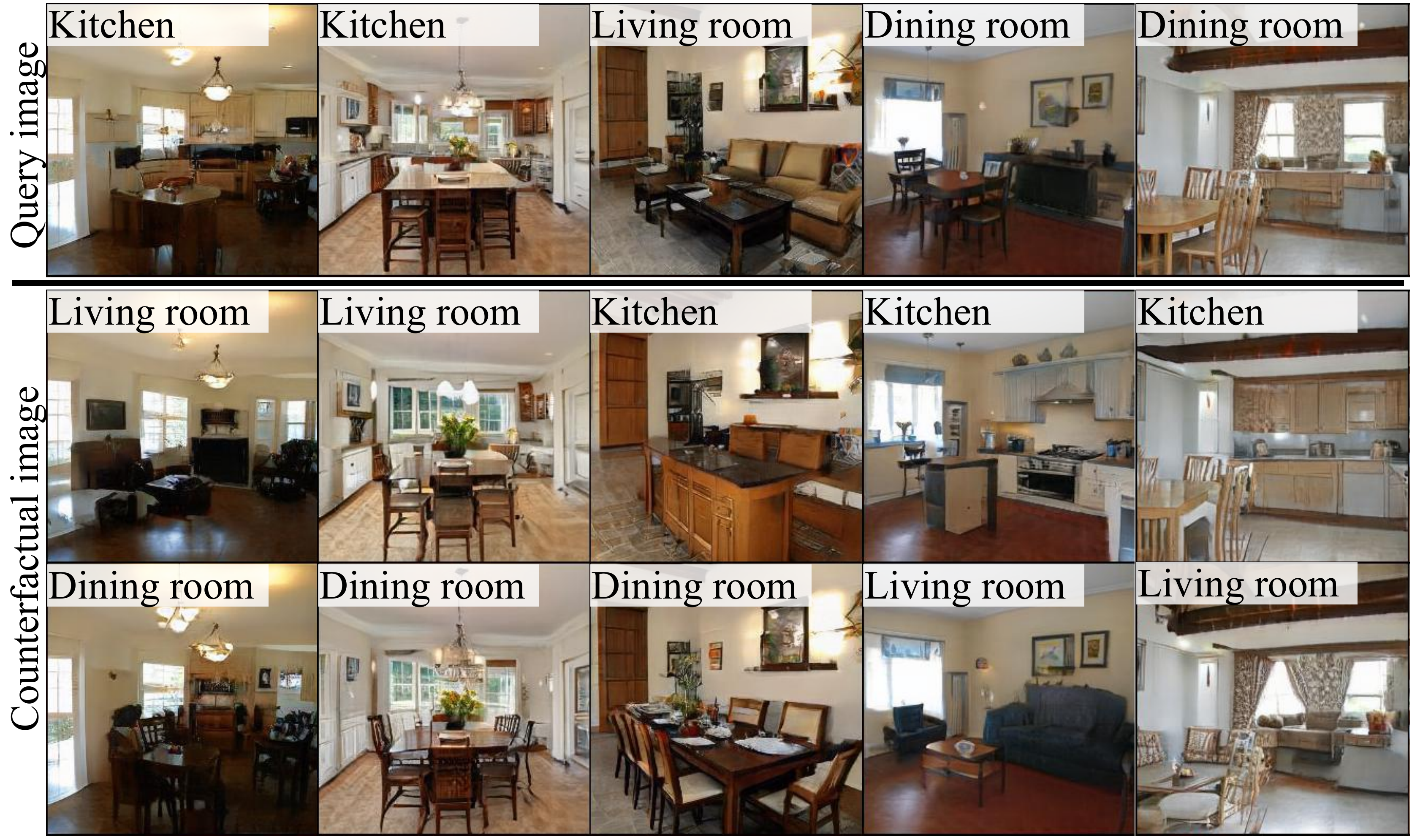}
    \caption{\label{fig:sup:lsun}\textbf{Qualitative results on LSUN dataset \cite{yu2015lsun}.}}
\end{figure*}

In \autoref{fig:sup:lsun}, we present preliminary results using the official pre-trained BlobGAN \cite{blobgan} generator on 3 classes of LSUN.
The decision model is a 3-class classifier trained on LSUN to distinguish between kitchens, living rooms and dining rooms.
We use generated images as queries.
We can observe for instance that while the presence of a sofa is a clear distinguishing feature for the decision model, chair style contributes as well (last column).
Also, while the explanations include layout changes impossible with STEEX \cite{steex}, the position of objects does not seem as important as in the BDD experiments as objects are not displaced as much.

\end{document}